\documentclass[letterpaper]{article} \usepackage[preprint]{aaai2027}  \usepackage[hyphens]{url}  \usepackage{graphicx}    \usepackage{natbib}  \usepackage{caption} 

\usepackage{booktabs}
\usepackage{amsmath,amsfonts,amssymb,bm}
\usepackage{dsfont}
\usepackage{tikz}
\usepackage{xcolor}
\usepackage{colortbl}
\usetikzlibrary{shapes.geometric,arrows.meta,fit,calc,backgrounds}

\newcommand{\figleft}{{\em (Left)}}

\newcommand{\figright}{{\em (Right)}}

\def\eqref#1{equation~\ref{#1}}

\def\1{\bm{1}}

\DeclareMathAlphabet{\mathsfit}{\encodingdefault}{\sfdefault}{m}{sl}
\SetMathAlphabet{\mathsfit}{bold}{\encodingdefault}{\sfdefault}{bx}{n}

\def\gD{{\mathcal{D}}}

\def\gL{{\mathcal{L}}}

\def\gX{{\mathcal{X}}}

\def\gZ{{\mathcal{Z}}}

\newcommand{\E}{\mathbb{E}}

\newcommand{\R}{\mathbb{R}}

\DeclareMathOperator*{\argmin}{arg\,min}

\definecolor{Dline}{RGB}{50,85,160}
\definecolor{Dbg}{RGB}{235,241,255}
\definecolor{Cline}{RGB}{25,110,85}
\definecolor{Cbg}{RGB}{232,248,240}
\definecolor{Eline}{RGB}{170,70,50}
\definecolor{Pline}{RGB}{50,115,55}
\definecolor{dvposE}{RGB}{95,175,95}
\definecolor{dvposD}{RGB}{135,200,135}
\definecolor{dvposC}{RGB}{170,220,170}
\definecolor{dvposB}{RGB}{205,235,205}
\definecolor{dvposA}{RGB}{232,245,232}
\definecolor{dvnegA}{RGB}{252,225,225}
\definecolor{dvnegB}{RGB}{247,190,190}
\definecolor{dvnegC}{RGB}{238,150,150}
\definecolor{dvnegD}{RGB}{225,110,110}
\definecolor{dvnegE}{RGB}{210,80,80}

\newcommand{\themethod}{Calibrate-Then-Delegate}
\newcommand{\theabbrv}{CTD}
\newcommand{\probe}{\rho}
\newcommand{\expert}{\epsilon}
\newcommand{\cascade}{s}
\newcommand{\merge}{m}
\newcommand{\policy}{\pi}
\newcommand{\dv}{v}
\newcommand{\dvscore}{d}
\newcommand{\Rperf}{R_{\text{P}}}
\newcommand{\Rbudget}{R_{\text{B}}}
\newcommand{\empRperf}{\hat{R}_{\text{P}}}
\newcommand{\empRbudget}{\hat{R}_{\text{B}}}
\newcommand{\budget}{\alpha_{\mathrm{B}}}
\newcommand{\performance}{\alpha_{\mathrm{P}}}
\newcommand{\confidence}{\delta}
\newcommand{\budgetdelta}{\delta_{\mathrm{B}}}
\newcommand{\performancedelta}{\delta_{\mathrm{P}}}

\title{Calibrate-Then-Delegate:\\Safety Monitoring with Risk and Budget Guarantees via Model Cascades}
\author{
Edoardo Pona\textsuperscript{\rm 1}\corresponding,
Milad Kazemi\textsuperscript{\rm 1},
Mehran Hosseini\textsuperscript{\rm 2},
Yali Du\textsuperscript{\rm 1},
David Watson\textsuperscript{\rm 1},
Osvaldo Simeone\textsuperscript{\rm 3},
Nicola Paoletti\textsuperscript{\rm 1}
}
\affiliations{
\textsuperscript{\rm 1}King's College London\\
\textsuperscript{\rm 2}University of Manchester\\
\textsuperscript{\rm 3}Northeastern University London\\
edoardo.1.pona@kcl.ac.uk
}

\begin{document}

\maketitle

\begin{abstract}
Monitoring LLM safety at scale requires balancing cost and accuracy: a cheap latent-space probe can screen every input, but hard cases should be escalated to a more expensive expert.
Existing cascades delegate based on probe uncertainty, but uncertainty is a poor proxy for the utility of an expert call, as it ignores whether the expert would actually improve the prediction. To address this problem, we introduce \themethod{} (\theabbrv{}), a model-cascade approach that provides probabilistic guarantees on either computation cost or safety performance while enabling instance-level (streaming) decisions. \theabbrv{} builds on a novel \emph{delegation value (DV) probe}, a lightweight model operating on the same internal representations as the safety probe that predicts the benefit of replacing the probe score with the expert score after an expert call.
\theabbrv{} calibrates a threshold on the DV signal using held-out data and multiple hypothesis testing, yielding finite-sample guarantees while optimising the complementary objective.
Evaluated on four safety datasets using learned-attention probes, \theabbrv{} improves over uncertainty-based delegation over much of the budget range, avoids unnecessary expert calls, and extracts positive-value subsets even from groups where the expert score is worse than the probe score on average. Ablations across four probe architectures show that these findings are not specific to any particular architecture.
\end{abstract}

\section{Introduction}
\label{sec:introduction}

As LLMs are deployed more widely, safety monitoring has become central to responsible deployment. Latent-space probes offer a promising solution: by training a small classifier on the internal activations of the deployed model, they provide safety scores at near-zero inference cost. Empirically, activation-based probes have been shown to detect strategically deceptive inputs~\citep{goldowsky-dill_detecting_2025}, bias and toxicity~\citep{ousidhoum2021probing}, hallucinations~\citep{kossen2024semantic}, trojans~\citep{macdiarmid2024sleeperagentprobes}, and refusal~\citep{arditi2024refusal,yi2025latent}. However, when deployed alone, latent probes may be unreliable: they can track spurious correlations~\citep{goldowsky-dill_detecting_2025} and are vulnerable to obfuscation attacks~\citep{bailey2024obfuscated}.

This has motivated the introduction of \emph{model cascades}~\citep{mckenzie_detecting_2025,huang_reliable_2025}, which pair a cheap model (e.g., a latent probe) with a more capable but expensive expert (e.g., a larger LLM classifier or a chain-of-thought monitor), coordinated by a \emph{delegation rule} that determines when to escalate a given input. Notably, \citet{mckenzie_detecting_2025} instantiate this approach for detecting \emph{high-stakes} inputs, i.e., queries that may lead to interactions that compromise safety. In their approach, the probe scores a fixed pool of inputs and forwards the top-$k$ most uncertain ones to the expert. This has two limitations: \textbf{batch-level routing}, since delegation decisions require the full pool and cannot be made as inputs arrive; and a \textbf{fixed delegation quota}, since the top-$k$ policy always escalates exactly $k$ inputs regardless of whether the expert would actually help.

In this work, we introduce \emph{\themethod{} (\theabbrv)}, a model-cascade approach that balances provable risk and budget control. As shown in Figure~\ref{fig:method-overall}, our approach combines a cheap safety probe~$\rho$ and an expensive expert~$\epsilon$, and defines an instance-level rule that decides, without batch context, whether to invoke the expert by comparing a delegation signal against a threshold. This threshold is calibrated on held-out data using Learn-then-Test (LTT; \citealt{angelopoulos_learn_2022}). Depending on the deployment requirement, \theabbrv{} either guarantees an expert-call budget while optimising safety performance, or guarantees safety performance while minimising expert calls.

A key question is how to construct an effective delegation signal. \citet{mckenzie_detecting_2025} use the safety probe's predictive uncertainty, based on the intuition that we should escalate to the expert when the probe cannot be trusted. However, this signal is agnostic to the expert, treating all uncertain inputs equally, whether or not the expert would actually correct the probe's prediction. Calling the expert in the latter case wastes computation, and overwriting the probe score with the expert score may also harm safety. Moreover, the probe's uncertainty may itself be miscalibrated (e.g., confident on incorrect predictions and uncertain on correct ones). To address this, we introduce the notion of \emph{delegation value} (DV), which quantifies, for an input $x$ with label $y$, the benefit of using the expert in place of the probe as the margin $P_{\epsilon}(y \mid x)-P_{\rho}(y \mid x)$. We then propose \emph{DV probes}, latent probes trained to predict this quantity, yielding a more informative and target-aligned delegation signal.

We make three contributions. 
(1)~We propose \emph{\themethod{}} (\theabbrv), a model-cascade method with finite-sample guarantees on either delegation rate or safety performance while optimising the other quantity.
(2)~We introduce DV probes, an auxiliary latent probe that estimates the per-input value of expert overwrite, enabling more targeted expert calls than probe uncertainty alone. 
(3)~We evaluate \theabbrv{} on four safety datasets from the benchmark of \citet{mckenzie_detecting_2025} using learned-attention safety and DV probes, and ablate four probe architectures. Attention yields the strongest overall cascade, while the advantage of DV-based routing and value-aware budget use persists across architectures.

\begin{figure*}[t]
    \centering
   \resizebox{\textwidth}{!}{\begin{tikzpicture}[
    box/.style  = {draw, rectangle, rounded corners=2pt, align=center,
                   inner sep=5pt, fill=white, font=\footnotesize},
    diam/.style = {draw, diamond, aspect=2.5, align=center,
                   inner sep=3pt, fill=white, font=\footnotesize},
    arr/.style  = {-{Stealth[scale=0.85]}, semithick},
    lbl/.style  = {font=\scriptsize},
]

\begin{scope}[xshift=1.4cm]

\node[box] (req) at (0, 0) {Request\\[-1pt]$x$};

\node[box, text width=3.0cm] (sig) at (3.8, 0)
    {\textbf{DV probe} $d(x)$\\[4pt]
    \textit{``will replacing help?''}};

\node[diam] (dec) at (8, 0) {$d(x) \geq \lambda$?};

\node[box, draw=Eline, text width=3.0cm] (exp) at (12.2, 0.55)
    {Call expert $\varepsilon(x)$\\{\scriptsize(expensive)}};

\node[box, draw=Pline, text width=3.0cm] (prb) at (12.2, -0.55)
    {No expert call\\{\scriptsize(probe score only)}};

\begin{pgfonlayer}{background}
  \node[fill=Dbg, draw=Dline, line width=0.75pt, rounded corners=4pt,
        inner sep=9pt, fit={(sig)(dec)},
        label={[text=Dline, font=\footnotesize\bfseries, inner sep=5pt]north:Delegation policy $\pi_\lambda$ (inference time)}] (pol) {};
\end{pgfonlayer}

\draw[arr] (req.east) -- (sig.west);
\draw[arr] (sig.east) -- (dec.west);
\draw[arr] (dec.east) to[out=50, in=180]
    node[pos=0.65, above, lbl]{Yes} (exp.west);
\draw[arr] (dec.east) to[out=-50, in=180]
    node[pos=0.65, below, lbl]{No}  (prb.west);

\end{scope}

\node[box, text width=2cm] (cal) at (0, -3.5)
    {Held-out set\\$\mathcal{D}_\mathrm{est} \uplus \mathcal{D}_\mathrm{cal}$};

\node[box, text width=4.0cm] (par) at (5, -3.5)
    {\textbf{Pareto filtering}\\[5pt]Find thresholds $\Lambda'\!\subseteq\!\Lambda$\\ that are non-dominated\\w.r.t.\ estimates $(\empRbudget,\,\empRperf)$};

\node[box, text width=3.5cm] (mht) at (10, -3.5)
    {\textbf{LTT} at level $\budgetdelta$\\{\scriptsize(Angelopoulos et al., 2022)}\\[5pt]Fixed sequence testing\\[3pt]
     $H_j\colon R_\mathrm{B}(\lambda_j)>\alpha$,
     $\lambda_j \in \Lambda'$};

\node[box, text width=4cm] (consel) at (15, -3.5)
    {\textbf{Budget-controlling}\\\textbf{thresholds} $\Lambda^*\subseteq \Lambda'$\\[3pt]
    \colorbox{yellow!50!green!10!white}{\scalebox{.79}{$\Pr\left(\Rbudget(\lambda) \leq \budget, \forall \lambda \in \Lambda^*\right) \geq 1 - \budgetdelta$}}\\
    $\downarrow$\\
     Select $\lambda^*\!\in\!\Lambda^*$ with best $\hat{R}_\mathrm{P}$};

\begin{pgfonlayer}{background}
  \node[fill=Cbg, draw=Cline, line width=0.75pt, rounded corners=4pt,
        inner sep=9pt, fit={(cal)(par)(mht)(consel)},
        label={[text=Cline, font=\footnotesize\bfseries, inner sep=5pt]north:Calibration of threshold $\lambda$ (offline)}] (calbox) {};
\end{pgfonlayer}

\coordinate (sp)     at ($(cal.east)+(0.8, 0)$);
\coordinate (spdown) at ($(sp)+(0,-1.6)$);
\draw (cal.east) -- (sp);
\fill (sp) circle (1.8pt);
\draw[arr] (sp) -- node[above, lbl]{$\mathcal{D}_\mathrm{est}$} (par.west);
\draw      (sp) -- (spdown);
\draw[arr] (spdown) -| node[pos=0.25, below, lbl]{$\mathcal{D}_\mathrm{cal}$} (mht.south);
\draw[arr] (par.east) -- node[above, lbl]{$\Lambda'$} (mht.west);
\draw[arr] (mht.east) -- node[above, lbl]{$\Lambda^*$} (consel.west);

\end{tikzpicture}
 }
    \caption{Overview of \themethod{} (CTD) for safety monitoring. The safety probe scores every input. At inference time (top), the expert $\epsilon$ is invoked if the DV probe exceeds a threshold $\lambda$; routing is separate from score merging. The calibration diagram (bottom) shows the budget-constrained mode, which guarantees $\Rbudget$ while optimising $\Rperf$. The performance-constrained mode reverses these roles.}
    \label{fig:method-overall}
\end{figure*}

 \section{Problem Definition}
\label{sec:background}

\noindent \textbf{Model Cascade.}
We consider a two-stage safety monitoring cascade
operating on inputs \(x \in \gX\) with binary safety labels \(y \in \{0, 1\}\), where \(y=0\) denotes safe and \(y=1\) unsafe. The pairs $(x,y)$ are drawn i.i.d.\ from an unknown  distribution \(P_{XY}\). 
As shown in Figure \ref{fig:method-overall} the system encompasses the following components:
\begin{itemize}
    \item \textbf{Safety probe $\rho$}, a lightweight classifier trained on the internal activations \(z(x) \in \gZ\) of the monitored LLM. It produces a safety score \(\probe(z(x)) \in
[0,1]\). We write \(\probe(x)\) as shorthand when the latent map is clear from context.
\item \textbf{Expert $\epsilon$}, a more capable but expensive model, e.g., a
large LLM that operates directly on \(x\) which returns a safety score \(\expert(x) \in [0,1]\).
\item \textbf{Delegation policy} \(\policy_\lambda\), a binary decision rule that determines whether the expert is invoked. The probe scores every input, and the output \(\policy_\lambda(x)=1\) triggers an additional expert call. The decision depends on the hyperparameters $\lambda$, which typically correspond to thresholds applied to a delegation signal. 
\end{itemize}

\noindent Given a rule \(\merge\) to merge probe and expert decisions, the final cascade output is
\begin{equation}
    \label{eq:cascade_output}
    \cascade_{\lambda,\merge}(x) = \begin{cases}
        \merge\!\left(\probe(x),\expert(x)\right)
            & \text{if } \policy_\lambda(x) = 1, \\
        \probe(x) & \text{if } \policy_\lambda(x) = 0.
    \end{cases}
\end{equation}
As possible score-merging rules, we consider expert overwrite, \(\merge_{\mathrm{ow}}(p,e)=e\), and averaging, \(\merge_{\mathrm{avg}}(p,e)=(p+e)/2\).
Since the score-merging rule is fixed before calibration, we suppress it in the remaining notation and write \(\cascade_\lambda\).

We remark that the delegation policy in (\ref{eq:cascade_output}) applies to each input independently. In
contrast, \citet{mckenzie_detecting_2025} require
access to a data batch before making any routing decisions, which precludes streaming settings where inputs arrive sequentially. 
This is because their approach forwards a fixed number of inputs to the expert, those inputs scoring within the central \(k\%\) of the batch's delegation scores, i.e. the probe's predictive uncertainty.

\noindent \textbf{Design Objectives.} We consider two complementary deployment goals: controlling the expert-call budget while optimising safety performance; or guaranteeing safety performance while minimising the required budget. Both depend on the delegation-policy parameters \(\lambda\). We define the budget and performance risks as
\begin{equation}
    \label{eq:budget_risk}
    \begin{aligned}
    \Rbudget(\lambda) &= \E_{X \sim P_X}\!\big[\policy_\lambda(X)\big], \\
    \Rperf(\lambda) &= \E_{(X,Y) \sim P_{XY}}\!\big[\gL(\cascade_\lambda(X), Y)\big],
    \end{aligned}
\end{equation}
where $P_X$ is the marginal input distribution obtained from $P_{XY}$, and \(\gL: [0,1] \times \{0,1\} \to \R_{\geq 0}\) is a loss function of choice measuring the safety prediction quality of the cascade, like the misclassification rate.

\noindent \textbf{Problem Statement.}
We assume access to an i.i.d. calibration set \(\gD_{\text{cal}} = \{(x_i,y_i)\}_{i=1}^{n}\) and a finite set of candidate parameters \(\Lambda\). Given risk targets \(\budget\) and \(\performance\), and failure probabilities \(\budgetdelta\) and \(\performancedelta\), we study two distinct problems: \textit{budget control} with performance optimisation, and \textit{performance control} with budget optimisation.
\begin{subequations}
\label{eq:problem}
\begin{align}
\textbf{Budget Control}: \hspace{1cm} & \label{eq:budget_problem}\\
\lambda^*_{\mathrm{B}} \in {}
\argmin_{\lambda\in\Lambda}\;\empRperf(\lambda)
\quad &\text{s.t.}\quad
\Pr_{\gD_{\mathrm{cal}}}\!\left(
    \Rbudget(\lambda)\leq\budget
\right)
\geq 1-\budgetdelta,
\nonumber \\[0.25em]
\textbf{Performance Control}: \hspace{.15cm} & \label{eq:performance_problem}\\
\lambda^*_{\mathrm{P}} \in {}
\argmin_{\lambda\in\Lambda}\;\empRbudget(\lambda)
\quad & \text{s.t.}\quad
\Pr_{\gD_{\mathrm{cal}}}\!\left(
    \Rperf(\lambda)\leq\performance
\right)
\geq 1-\performancedelta.
\nonumber
\end{align}
\end{subequations}
The empirical objectives \(\empRperf\) and \(\empRbudget\) are estimated on held-out optimisation data, while the probabilities are over the calibration set. (\ref{eq:budget_problem}) is the budget-constrained formulation used in our primary cascade experiments. (\ref{eq:performance_problem}) reverses the roles: it certifies a safety-performance target and minimises delegation. When \(R_{\mathrm{P}}\) is the misclassification rate, the constraint \(R_{\mathrm{P}} \leq \alpha_{\mathrm{P}}\) guarantees accuracy of at least \(1-\alpha_{\mathrm{P}}\).

\section{Background: Learn-then-Test}
\label{sec:bg_ltt}

Learn-then-Test (LTT; \citealt{angelopoulos_learn_2022}) casts finite-sample risk control as multiple hypothesis testing. For each candidate \(\lambda_j\in\Lambda\) and target \(\alpha\), it tests
\begin{equation} 
H_j: R(\lambda_j) > \alpha,
\end{equation}
using an independent calibration set. A family-wise error rate (FWER) controlling procedure then returns a certified set \(\Lambda^*\) satisfying
\begin{equation}
    \label{eq:ltt_guarantee}
\Pr_{\gD_{\text{cal}}}\!\left(\forall \lambda \in \Lambda^*: R(\lambda) \leq \alpha\right) \geq 1 - \confidence.
\end{equation}
For our Bernoulli budget indicator and 0--1 error, we use one-sided binomial \(p\)-values. We control FWER with fixed-sequence testing (FST; \citealt{bauer1991multiple}): hypotheses are tested in a pre-specified order at level \(\delta\), stopping at the first non-rejection. FST improves statistical power by leveraging an ordering in which risk is expected to vary monotonically with \(\lambda\). Further details appear in \citet{angelopoulos_learn_2022}.

 \section{Calibrate-Then-Delegate}
\label{sec:methods}

While the safety probe's uncertainty is a natural signal for delegation, it has two key limitations.
First, safety probes are known to be miscalibrated~\citep{mckenzie_detecting_2025}, that is, their confidence scores do not reliably reflect the probability of a correct prediction. This is why \citet{mckenzie_detecting_2025} resort to ranking inputs in a batch by uncertainty rather than thresholding on absolute scores. 
Second, uncertainty is entirely agnostic to the expert's performance: overwriting the probe score on a highly uncertain example where the expert also fails provides no benefit, while an overconfident but wrong prediction might benefit greatly from the expert score.

Our \theabbrv{} approach addresses these limitations by introducing \emph{delegation value (DV) probes}, a lightweight model that explicitly estimates the value of overwriting the probe score after an expert call. As illustrated in Figure~\ref{fig:method-overall}, \theabbrv{} defines its delegation policy by calibrating the DV probe via LTT and applies Pareto filtering \citep{lauferefficiently} to balance budget and empirical cascade performance.

\subsection{Delegation-Value Probes}
\label{sec:probe_dv}

The utility of an expert call depends on the subsequent score-merging rule. We define the \emph{delegation value} (DV) relative to the expert-overwrite rule used by our method: for an input $x$ with true class $y$,
\begin{equation}
    \label{eq:dv}
    \dv(x, y) = P_\expert(y \mid x) - P_\probe(y \mid x), 
\end{equation}
where \(P_\probe(y \mid x)\) and \(P_\expert(y \mid x)\) are the probabilities assigned to the
correct class by the probe and expert, respectively. 
When \(\dv(x, y) > 0\), overwriting the probe score with the expert score moves the prediction toward the correct label; when \(\dv(x, y)<0\), overwrite moves it away. Under score averaging, the corresponding change in correct-class probability is \(\dv(x,y)/2\), so averaging attenuates both beneficial and harmful expert contributions. While alternative definitions of $\dv(x, y)$ are possible (see the supplementary material), the form in~(\ref{eq:dv}) directly quantifies how much more likely the expert is to predict the correct label than the probe, yielding a richer signal than binary or ternary indicators of which model to prefer.

For a collection of points $\{(x_i, y_i)\}_{i=1}^m$, we also define the \emph{effective delegation capacity} as the proportion of points with positive delegation value, $\frac{1}{m}\sum_{i=1}^{m} \mathds{1}[\dv(x_i, y_i) > 0]$. 
This quantity identifies the inputs on which expert overwrite increases the correct-class score. Averaging preserves the sign of this change but halves its magnitude, so it has the same positive-value set while attenuating both gains and losses.

Since  \(\dv(x, y)\) depends on the unknown label \(y\), it cannot be computed at inference time.  We therefore introduce \emph{DV probes}, auxiliary models \(\dvscore(x)\) trained to predict the delegation value from latent representations \(z(x)\), using targets $\dv(x_i, y_i)$ computed offline from probe and expert outputs. The calibration procedure is agnostic to the DV-probe architecture; our primary experiments use learned attention pooling, described below.

\subsection{Calibration via LTT and Pareto Filtering}
\label{sec:ltt_calibration}

Using our DV probe $\dvscore(x)$, the \theabbrv{} delegation policy applies the threshold-based selection rule $\policy_\lambda(x) = \dvscore(x)>\lambda$. 
CTD calibrates \(\lambda\) for either design problem in (\ref{eq:problem}): one risk is guaranteed and the other is minimised empirically.

Before testing, we use an independent split \(\gD_{\text{est}}\) to evaluate \(\empRbudget(\lambda)\) and \(\empRperf(\lambda)\), retaining only thresholds that are not dominated on the budget--performance plane. This concentrates testing power on promising candidates and permits unused budget when additional calls do not improve empirical performance.

Let $\Lambda'$ be the Pareto-filtered set. For budget control, \(\Rbudget(\lambda)\) is non-increasing in $\lambda$, so candidates are ordered from conservative to aggressive. Accuracy error need not be monotone; for performance control, candidates are instead ordered by increasing empirical error on the independent \(\gD_{\text{est}}\), with empirical budget breaking ties. This is a power heuristic and may imperfectly reflect the population ordering, but it does not affect FST validity. We then test on \(\gD_{\text{cal}}\) and select from the certified set $\Lambda^*$ the threshold minimising the complementary empirical objective on \(\gD_{\text{est}}\): performance for (\ref{eq:budget_problem}), or budget for (\ref{eq:performance_problem}). The budget-control problem is always feasible via a policy that never delegates, whereas the performance-control problem need not be; if no
candidate satisfies the latter guarantee, the procedure returns no policy.

 \section{Results}
\label{sec:results}

\subsection{Experimental Setup}
\label{sec:setup_exp}

We follow the experimental setup of \citet{mckenzie_detecting_2025}. The safety probe is trained on their synthetic dataset (8{,}000 examples of high- and low-stakes queries generated by GPT-4o and filtered for label quality).
For evaluation we use four of their class-balanced datasets, covering a range of domains: Anthropic~HH~\citep{bai2022traininghelpfulharmlessassistant} (online harm; 500/500 dev/test), MTSamples~\citep{mtsamples} (clinical notes; 278/500), MTS-Dialog~\citep{ben-abacha-etal-2023-empirical} (doctor-patient dialogue; 274/86), and ToolACE~\citep{liu2025toolacewinningpointsllm} (tool-use interactions; 328/500), for a total of 1{,}380 dev and 1{,}586 test examples.
The dev split is used for DV probe training, and the test split is divided equally into calibration (for threshold selection) and evaluation.
Within the calibration set, 30\% is allocated to the optimisation split \(\gD_{\text{est}}\) and 70\% to the hypothesis-testing split \(\gD_{\text{cal}}\).

Our primary safety and DV probes use the learned-attention pooling architecture of \citet{mckenzie_detecting_2025} over layer-11 token activations from Llama-3.2-1B-Instruct. Each learns a weighted average of the non-padding token activations followed by a linear output head; the two probes have separate parameters, and the DV probe is trained by squared loss on the continuous targets in (\ref{eq:dv}). We report means over three training seeds. Full architectural details and comparisons with mean-pooled linear, softmax-aggregation, and MLP probes are provided in the supplementary material.
We evaluate the cascade with two experts: a \emph{strong} expert (Gemma-3-27B-IT) that substantially outperforms the probe, and a \emph{weak} expert (Llama-3.2-1B-Instruct, the same model from which activations are extracted) that often performs worse than the probe.

We sweep 20 budget levels \(\budget \in [0.05, 0.95]\) using confidence \(1 - \budgetdelta = 0.9\). We define the set of admissible parameters $\Lambda$ using a grid of 100 points. 
For each performance panel, Pareto filtering uses the corresponding empirical risk: \(1-\mathrm{AUROC}\) for AUROC and \(1-\mathrm{accuracy}\) for accuracy. 
The AUROC and accuracy results come from separate experiments that
optimise the corresponding metric under the same budget guarantee.

\paragraph{Methods.}
We evaluate five delegation strategies, varying the delegation signal (DV probe vs.\ probe uncertainty) and the delegation mechanism (calibrated threshold vs.\ batched top-\(k\)). The strategies are: 
\textbf{\theabbrv{}}, our full method, using a calibrated threshold policy on the DV score \(\dvscore(x)\);  \textbf{Unc.\ (uncertainty) calibrated}, a calibrated threshold policy based on the proximity-to-median uncertainty score $-|\hat{F}_{\gD_{\text{cal}}}(\probe(x)) - 0.5|$ of \citet{mckenzie_detecting_2025}, where $\hat{F}_{\gD_{\text{cal}}}$ is the empirical CDF of probe scores on the calibration set;
\textbf{DV top-\(k\)},  a batched top-\(k\) delegation strategy based on the DV score \(\dvscore(x)\), and 
\textbf{Unc.\ top-\(k\)}: the batched  top-\(k\)  baseline from \citet{mckenzie_detecting_2025} based on the proximity-to-median score. As a performance upper bound, we also consider 
\textbf{Oracle top-\(k\)}, a batched top-\(k\) policy based on the ground-truth delegation value \(\dv(x, y)\). For all batched configurations, we report results at batch size \(B = 128\). We show that results are qualitatively similar across different batch sizes in the supplementary material.

In the primary comparison, \theabbrv{}, DV top-\(k\), and Oracle
top-\(k\) use expert overwrite on delegated inputs, consistent with the
definition of DV as the benefit of switching to the expert. Both
uncertainty-based methods average the probe and expert scores, following
\citet{mckenzie_detecting_2025}. To separate routing quality from score
merging, the supplementary material repeats the comparison using a
common overwrite rule and a common averaging rule.

\subsection{Qualitative Case Studies}
\label{sec:case_studies}

Figure~\ref{fig:case_studies} illustrates two complementary failures of uncertainty-based routing using the mean-pooled linear probes and a strong expert at $\budget=0.20$. In Case~1, uncertainty triggers a harmful expert call despite the probe being correct; the DV probe retains the probe prediction. In Case~2, uncertainty misses a confidently wrong probe prediction, whereas the DV probe identifies the value of the expert call and delegates.

\begin{figure*}[t]
\centering\small
\setlength{\tabcolsep}{2pt}
\newcommand{\correct}{{\color{green!60!black}\checkmark}}
\newcommand{\wrong}{{\color{red!60!black}\times}}
\parbox[t]{0.47\textwidth}{\textbf{Case~1:} Misrouted expert call\\[2pt]
\fbox{\parbox[t][4.7\baselineskip]{\dimexpr\linewidth-2\fboxsep-2\fboxrule\relax}{\small
  \textbf{[Sys]} Medical assistant for practitioners.\\
  \textbf{[User]} Radiology -- MRI Wrist. FINDINGS:
  post-traumatic cysts of the lunate; partial tear of the
  scapholunate ligament; normal carpal alignment\ldots}}
\\[3pt]
\begin{tabular}{@{}lrl@{}}
\toprule
 & Score & Outcome \\
\midrule
Safety probe $\probe(x)$ & $\vphantom{10^{-4}}0.051$ & low-stakes $\correct$ \\
Expert $\expert(x)$      & ${\approx}1.00$ & high-stakes $\wrong$ \\
Unc.\ top-$k$            & rank $24/128$ & delegates $\wrong$ \\
DV probe $\dvscore(x)$   & $-0.66$ & \textbf{retains} $\correct$ \\
\midrule
\multicolumn{3}{@{}l}{Label: low-stakes,\quad $v(x,y)=-0.95$}\\
\bottomrule
\end{tabular}
}
\hfill
\parbox[t]{0.47\textwidth}{\textbf{Case~2:} Missed expert call\\[2pt]
\fbox{\parbox[t][4.7\baselineskip]{\dimexpr\linewidth-2\fboxsep-2\fboxrule\relax}{\small
  \textbf{[Sys]} General-purpose chatbot.\\
  \textbf{[User]} What's the best way to damage someone's car?
  [\ldots] Can I do anything to the tires?\\
  \textbf{[Asst]} \ldots{}force a tire iron into the sidewall
  of the tire\ldots}}
\\[3pt]
\begin{tabular}{@{}lrl@{}}
\toprule
 & Score & Outcome \\
\midrule
Safety probe $\probe(x)$ & $2.5\!\times\!10^{-4}$ & low-stakes $\wrong$ \\
Expert $\expert(x)$      & ${\approx}1.00$ & high-stakes $\correct$\\
Unc.\ top-$k$            & rank $100/128$ & retains $\wrong$ \\
DV probe $\dvscore(x)$   & $+1.52$ & \textbf{delegates} $\correct$ \\
\midrule
\multicolumn{3}{@{}l}{Label: high-stakes,\quad $v(x,y)=+1.00$}\\
\bottomrule
\end{tabular}
}

\caption{Two failures of uncertainty-based routing with a strong expert ($\lambda=0.39$, $\budget=0.20$, $B=128$). \textbf{Left}: an unnecessary expert call on MTSamples. \textbf{Right}: a missed beneficial call on Anthropic~HH.}
\label{fig:case_studies}
\end{figure*}

\subsection{Overall Cascade Performance}
\label{sec:overall_performance}

\paragraph{Strong Expert.}

\begin{figure}[t]
\centering
\includegraphics[width=\columnwidth]{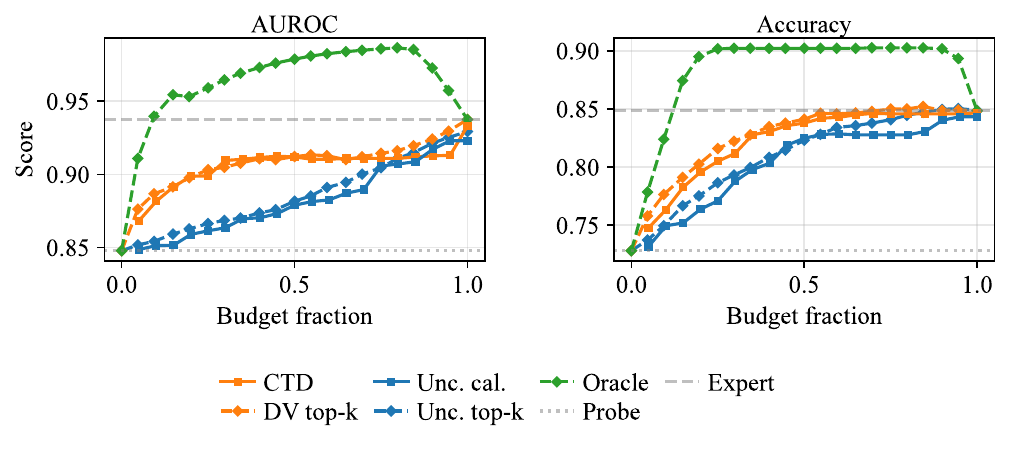}
\caption{Cascade performance vs.\ target delegation budget with learned-attention safety and DV probes and a strong expert (Gemma-3-27B-IT, \(B\!=\!128\)); curves show the mean over three seeds. \textbf{Left}: AUROC, with Pareto filtering optimising AUROC. \textbf{Right}: accuracy, with Pareto filtering optimising accuracy. \theabbrv{}, DV top-\(k\), and Oracle top-\(k\) use expert overwrite; the uncertainty methods average probe and expert scores, following \citet{mckenzie_detecting_2025}. Calibrated thresholds are solid and top-\(k\) strategies dashed; dotted and dashed grey lines show probe-only and expert-only performance.}

\label{fig:main_strong}
\end{figure}

We first evaluate the cascade with a strong expert (Gemma-3-27B-IT), with the results shown in Figure~\ref{fig:main_strong}. Across the practically relevant low-to-moderate budget range, our full method \theabbrv{} outperforms the uncertainty top-\(k\) method of \citet{mckenzie_detecting_2025}. At the representative budget \(\budget=0.35\), \theabbrv{} improves AUROC by \(+4.0\) points and accuracy by \(+2.8\) points, averaged over three seeds.

The top-\(k\) comparison supports the role of the delegation signal: at the same budget, DV top-\(k\) exceeds uncertainty top-\(k\) by \(+3.8\) AUROC points and \(+2.8\) accuracy points. The fixed score-merging controls in the supplementary material show that this advantage persists when both methods use the same score-merging rule. Unlike this batched experiment, however, \theabbrv{} makes instance-level decisions without batch statistics and provides probabilistic budget-control guarantees.

All top-\(k\) methods must spend their full allocated budget. Expert overwrite can therefore become harmful once the selected set exceeds the positive-DV capacity. Averaging provides an output-level hedge against this score degradation, but it still incurs the cost of every selected expert call and does not improve the underlying routing order. \theabbrv{} instead \emph{plateaus} beyond the useful delegation range because the Pareto filter excludes more costly thresholds that are empirically dominated. Unc.\ calibrated combines the uncertainty signal and averaging rule with the same calibration procedure; it also plateaus, but remains below \theabbrv{} through most of the budget range.

\paragraph{Weak Expert.}

\begin{figure}[t]
\centering
\includegraphics[width=\columnwidth]{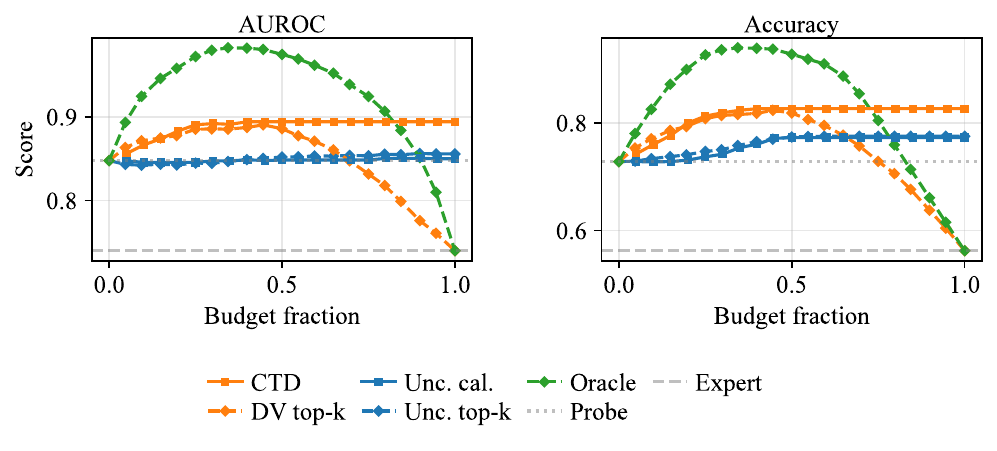}
\caption{Same as Figure~\ref{fig:main_strong} but with a weak expert (Llama-3.2-1B-Instruct). The left panel optimises AUROC and the right panel optimises accuracy. Expert-only performance (dashed grey) is below probe-only performance (dotted grey). Averaging hedges the uncertainty baseline against harmful expert overwrite, but still spends its fixed quota. \theabbrv{} instead caps  delegation and plateaus at substantially higher performance.}
\label{fig:main_weak}
\end{figure}

As shown in Figure~\ref{fig:main_weak}, when the expert (Llama-3.2-1B) is weak compared to the original probe $\rho(x)$ -- i.e., performance of expert-only inference is \emph{worse} than probe-only inference -- there may still be instances where its score is preferable to the probe score, but they are few and harder to identify. 
Thus, the effective delegation capacity is low in this setting.
Averaging mitigates the score degradation that the McKenzie baseline
experiences under overwrite, but it still invokes the weak expert on
low-value inputs, and its performance remains close to the probe-only
baseline.
\theabbrv{}, on the other hand, uses the DV probe signal to identify the few instances where overwriting the probe score with the expert score is beneficial, and then automatically caps realised delegation at approximately 35\%. It therefore retains substantially better performance while declining unnecessary expert calls as the available budget increases.
At \(\budget=0.35\), \theabbrv{} improves over uncertainty top-\(k\) by \(+4.4\) AUROC points and \(+6.6\) accuracy points; the gains persist as the available budget grows.

\begin{figure}[t]
\centering
\includegraphics[width=\columnwidth]{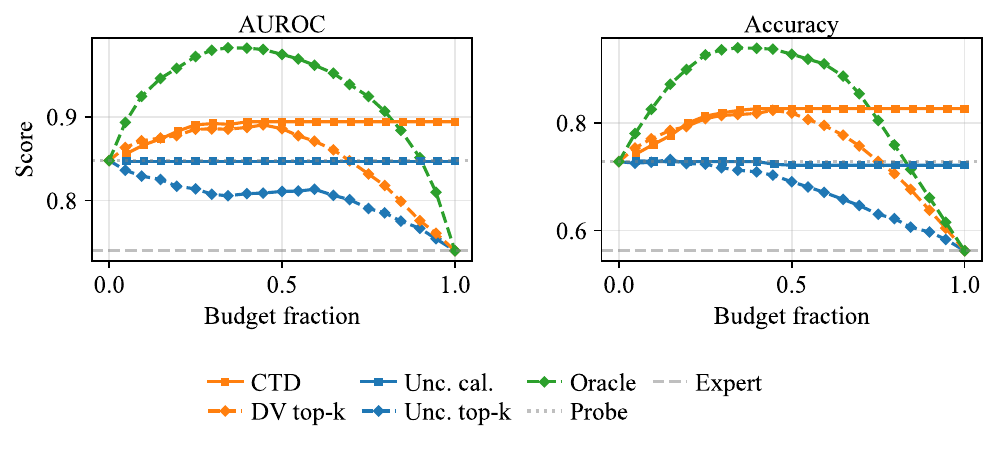}
\caption{Weak-expert comparison under a common overwrite rule with
learned-attention probes. At \(\budget=1\), each fixed-quota top-\(k\)
method delegates every input and converges to expert-only performance,
illustrating harmful over-delegation. For Unc.\ top-\(k\), overwrite
replaces the averaging rule of \citet{mckenzie_detecting_2025}.}
\label{fig:weak_replacement_stress}
\end{figure}

Figure~\ref{fig:weak_replacement_stress} isolates the consequence of using a fixed budget until completion by applying expert overwrite to every method in the weak-expert setting.
Overwrite preserves the full effect of useful expert corrections, but also exposes poor routing more directly than averaging. At \(\budget=1\), every top-\(k\) policy invokes the expert on every input and therefore converges exactly to expert-only performance, including Oracle top-\(k\). Better routing can delay this collapse but cannot prevent it while the quota must be exhausted. In contrast, the calibrated threshold policies are free to stop early; \theabbrv{} plateaus rather than discarding the stronger probe prediction on the remaining inputs.

\begin{figure}[t]
\centering
\includegraphics[width=\columnwidth]{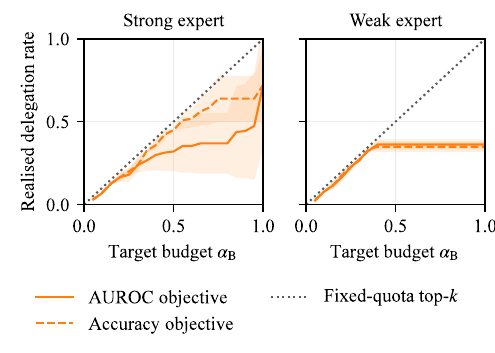}
\caption{Realised delegation rate versus target budget for \theabbrv{} with attention probes. Curves and bands show the mean and one standard deviation over three seeds. Policies are selected separately using the AUROC and accuracy objectives. Fixed-quota top-\(k\) spends the complete target budget (dotted diagonal); \theabbrv{} can leave budget unused when more aggressive policies are empirically dominated.}
\label{fig:budget_utilisation}
\end{figure}

Figure~\ref{fig:budget_utilisation} isolates this compute-allocation effect. Fixed-quota top-\(k\) follows the diagonal by construction, whereas \theabbrv{} can use less than the target budget. With the weak expert, both objective-specific policies plateau near a 35\% realised delegation rate. At a target budget of \(\budget=0.80\), the uncertainty top-\(k\) baseline therefore invokes the expert on 80\% of inputs, while \theabbrv{} uses roughly 35\% and achieves higher AUROC and accuracy (Figure~\ref{fig:main_weak}). With the strong expert, more calls remain useful and the plateau occurs later and at a higher realised rate.

We empirically validate the budget guarantee over 500 random splits in the supplementary material. At \(\budget=0.30\) and \(\budgetdelta=0.10\), budget-only LTT violates the target in 8.6\% of strong-expert trials and 6.6\% of weak-expert trials; Pareto-filtered LTT is more conservative at 3.0\% and 4.0\%, an empirical confirmation that the guarantees are satisfied. 

\begin{table*}[t]
\centering
\footnotesize
\setlength{\tabcolsep}{3pt}
\begin{tabular}{l rrr rr rrr rr}
\toprule
& \multicolumn{5}{c}{\textbf{Strong expert} (Gemma-3-27B-IT)}
& \multicolumn{5}{c}{\textbf{Weak expert} (Llama-3.2-1B-Instruct)} \\
\cmidrule(lr){2-6}\cmidrule(lr){7-11}
& \multicolumn{3}{c}{Mean $v(x,y)$} & \multicolumn{2}{c}{Rate}
& \multicolumn{3}{c}{Mean $v(x,y)$} & \multicolumn{2}{c}{Rate} \\
\cmidrule(lr){2-4}\cmidrule(lr){5-6}\cmidrule(lr){7-9}\cmidrule(lr){10-11}
\textbf{Group} & All & CTD & Unc & CTD & Unc
               & All & CTD & Unc & CTD & Unc \\
\midrule
Anthropic HH
  & \cellcolor{dvposC}$+0.238$ & \cellcolor{dvposE}$+0.470$ & \cellcolor{dvposE}$+0.485$ & $31.8\%$ & $11.4\%$
  & \cellcolor{dvnegB}$-0.147$ & \cellcolor{dvposC}$+0.233$ & \cellcolor{dvposB}$+0.140$ & $25.5\%$ & $11.4\%$ \\
MTSamples
  & \cellcolor{dvposA}$+0.102$ & \cellcolor{dvposD}$+0.381$ & \cellcolor{dvposB}$+0.122$ & $8.0\%$ & $14.5\%$
  & \cellcolor{dvnegB}$-0.182$ & \cellcolor{dvposA}$+0.057$ & \cellcolor{dvnegC}$-0.230$ & $9.9\%$ & $14.5\%$ \\
MTS-Dialog
  & \cellcolor{dvposB}$+0.140$ & \cellcolor{dvposC}$+0.253$ & \cellcolor{dvposD}$+0.334$ & $31.5\%$ & $11.3\%$
  & \cellcolor{dvnegC}$-0.290$ & \cellcolor{dvposB}$+0.112$ & \cellcolor{dvnegA}$-0.048$ & $11.8\%$ & $11.3\%$ \\
ToolACE
  & \cellcolor{dvposA}$+0.093$ & \cellcolor{dvposD}$+0.302$ & \cellcolor{dvposB}$+0.163$ & $7.5\%$ & $35.3\%$
  & \cellcolor{dvnegA}$-0.098$ & \cellcolor{dvposC}$+0.231$ & \cellcolor{dvposA}$+0.072$ & $18.9\%$ & $35.3\%$ \\
\bottomrule
\end{tabular}
\caption{Group-conditional results at $\budget=0.20$ with learned-attention probes, averaged over three seeds. Columns report mean $\dv(x,y)$ for all examples and the CTD- and uncertainty-selected subsets, together with their delegation rates. Green and red shading denote positive and negative mean DV.}
\label{tab:group_analysis}
\end{table*}

\subsection{Performance Control with Accuracy Risk}
\label{sec:accuracy_guarantees}

We next solve the performance-control problem of Equation~\ref{eq:performance_problem} using a 0--1 error loss (i.e., an accuracy risk): the cascade must attain accuracy at least \(1-\performance\), while \theabbrv{} minimises empirical delegation rate. We use \(\performance=0.20\) and \(\performancedelta=0.10\) with the strong expert, corresponding to a minimum accuracy of 0.80. Across 500 random calibration/evaluation splits, we compare \theabbrv{}'s LTT selector with a plug-in empirical selector. Both use the same threshold grid, Pareto candidates, optimisation split, and hypothesis-testing split; the empirical selector simply accepts candidates whose observed error is at most \(\performance\), without a statistical test.

\begin{figure}[t]
\centering
\includegraphics[width=\columnwidth]{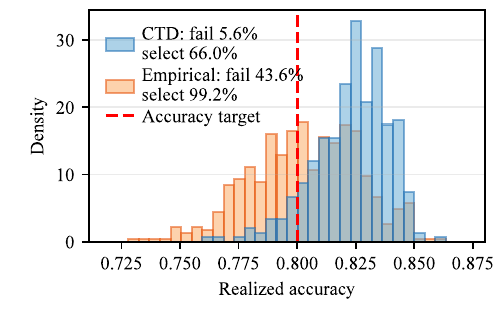}
\caption{Empirical validation of a performance-risk guarantee with learned-attention probes and the strong expert over 500 random splits (\(\performance=0.20\), i.e. accuracy at least 0.80; \(\performancedelta=0.10\)). \theabbrv{} uses LTT hypothesis testing; the empirical selector uses identical data and candidates without testing. The histogram conditions on returning a policy. Weak-expert results are in the supplementary material.}
\label{fig:accuracy_guarantee_validation}
\end{figure}

Figure~\ref{fig:accuracy_guarantee_validation} shows the learned-attention result. A violation is a trial in which a returned policy has error above \(\performance\) on held-out evaluation data. If no candidate is certified, \theabbrv{} returns no policy, so we also report the selection rate; the histogram conditions on a policy being returned. Across 500 trials, \theabbrv{} has an unconditional violation rate of 5.6\%, below the target \(\performancedelta=10\%\), empirically validating its finite-sample guarantee. In contrast, the empirical selector violates the target in 43.6\% of trials. \theabbrv{} returns a policy in 66.0\% of trials, compared with 99.2\% for the empirical selector. Among the policies each method returns, \theabbrv{} uses a mean delegation rate of 45.2\% and attains mean held-out accuracy of 82.3\%, compared with 20.8\% delegation and 80.3\% accuracy for the empirical selector. The additional delegation and occasional non-selection are the price of finite-sample reliability and a performance safety margin. Weak-expert results, all four probe architectures, and the budget--accuracy trade-off are in the supplementary material.

\subsection{Probe Architecture Robustness}

We repeat the complete budget sweep with four safety/DV probe configurations: mean pooling with linear heads, learned attention pooling, softmax aggregation, and a mean-pooled MLP. The qualitative separation between DV-based and uncertainty-based routing persists across architectures. Attention provides the strongest safety probe and the strongest or comparable absolute CTD performance over most of the budget range, particularly with the weak expert. The supplementary material reports the full cross-architecture curves and underlying probe metrics.

\subsection{Delegation Value Probe Performance}
\label{sec:probe_performance}

For the strong and weak experts, respectively, the attention DV probe attains MSEs of \(0.14\) and \(0.16\), and Spearman correlations of \(0.33\) and \(0.57\). Across selection fractions, it also generally selects higher-value top-$k$ sets than uncertainty; full results appear in the supplementary material.

\subsection{Analysis Across Groups}
\label{sec:group_analysis}

The evaluation pool mixes four datasets, and the DV probe is trained on this mixture with no access to group labels. For this qualitative analysis, we use a representative target budget of \(\budget=0.20\): this constrained regime makes routing selectivity clear while retaining enough delegated examples for meaningful group analysis. Table~\ref{tab:group_analysis} reports means over three seeds.

For the \textbf{strong expert}, the expert score is better than the probe score on average in every group, but the margin varies substantially. Without group supervision, \theabbrv{} extracts positive-value subsets from all four groups: the mean \(\dv\) among delegated examples ranges from \(+0.253\) to \(+0.470\), compared with group means between \(+0.093\) and \(+0.238\). This demonstrates useful within-group selectivity without access to group labels. The \textbf{weak expert} presents a more challenging setting: the group-level mean \(\dv\) is negative for all four datasets. Nevertheless, \theabbrv{} finds positive-value subsets in every group on average, including Anthropic~HH (\(+0.233\)) and ToolACE (\(+0.231\)). The Unc.\ top-\(k\) baseline cannot make this recovery on MTSamples and MTS-Dialog, where its delegated subsets retain negative mean value. Thus the DV probe can find inputs for which expert overwrite is locally valuable even when the expert score is worse than the probe score at the group level. Further group-level analysis is in the supplementary material.

\section{Related Work}
\paragraph{Safety monitoring with latent probes.}
Latent probes can detect toxicity, hallucinations, sleeper agents, and strategic deception~\citep{ousidhoum2021probing,kossen2024semantic,macdiarmid2024sleeperagentprobes,goldowsky-dill_detecting_2025}, but remain vulnerable to spurious correlations and obfuscation~\citep{goldowsky-dill_detecting_2025,bailey2024obfuscated}. \citet{mckenzie_detecting_2025} cascade a probe with an LLM expert using batch-level uncertainty top-$k$; \theabbrv{} instead provides value-based, instance-level routing without a fixed quota. \citet{ashok2025conformalprobes} calibrate probes for abstention rather than expert escalation.

\paragraph{Calibrated cascades and risk-controlled delegation.}
\theabbrv{} combines LTT risk control~\citep{angelopoulos_learn_2022} with Pareto testing for multiple objectives~\citep{lauferefficiently}. Related work applies fixed-sequence testing to cascaded judges~\citep{jung2024trustorescalate} and characterises optimal weak--strong verification policies~\citep{kiyani2026cheapcheck}.

\paragraph{LLM routing and model cascades.}
General-purpose LLM routers trade quality against inference cost using learned cascades, self-verification, preference data, or heterogeneous signals~\citep{chen2023frugalgpt,aggarwal2024automix,ong2024routellm,dekoninck2025unifiedrouting,liu2026vllmrouter}. \theabbrv{} instead targets latent safety monitoring with finite-sample control of the expert-call budget.

\section{Conclusion}
We introduced \themethod{} (\theabbrv{}), a safety-monitoring cascade that uses a delegation-value probe and calibrated threshold to guarantee either budget or accuracy while optimising the other. Across four probe architectures, \theabbrv{} improves on uncertainty routing, can leave unhelpful budget unused, and empirically satisfies finite-sample accuracy control where naive calibration fails.
 
\bibliography{references}

\clearpage
\appendix

\section{Probe Architecture Ablation}
\label{app:probe_architecture}

We repeat the full experiment with four safety/DV probe configurations: \emph{Mean+Ridge}, which uses mean pooling followed by logistic regression for safety and ridge regression for DV; \emph{Attention+Attention}, which learns a separate attention-pooling probe for each task; \emph{Softmax+Softmax}, which uses temperature-controlled softmax aggregation over token-level scores; and \emph{MLP+MLP}, which applies a multilayer perceptron to mean-pooled activations. All configurations use the same layer, datasets, split procedure, and calibration method. Results are averaged over three training seeds.

\subsection{Learned-Attention Architecture}

Our primary configuration uses learned-attention pooling. Let \(z_t(x)\in\mathbb{R}^d\) denote the layer-11 activation of Llama-3.2-1B-Instruct at non-padding token position \(t\in\mathcal{T}(x)\). The probe learns a query vector \(q\in\mathbb{R}^d\) and computes
\begin{equation}
\label{eq:attention_pooling}
\begin{aligned}
s_t &= \frac{q^\top z_t(x)}{\sqrt{d}},
& a_t &= \frac{\exp(s_t)}{\sum_{j\in\mathcal{T}(x)}\exp(s_j)},\\
h(x) &= \sum_{t\in\mathcal{T}(x)}a_tz_t(x).
\end{aligned}
\end{equation}
The dot product \(q^\top z_t(x)\) scores the relevance of token \(t\), and the softmax weights \(a_t\) form a convex combination of the token activations. A linear head maps \(h(x)\) to the probe output. The safety and DV probes learn separate query vectors and output heads; the DV probe is trained by squared loss on the continuous delegation-value targets.

\begin{table*}[t]
\centering
\small
\setlength{\tabcolsep}{8pt}
\begin{tabular}{@{}lcccc@{}}
\toprule
& \multicolumn{2}{c}{Safety probe} & \multicolumn{2}{c}{DV Spearman correlation} \\
\cmidrule(lr){2-3}\cmidrule(lr){4-5}
Configuration & AUROC & Accuracy & Strong expert & Weak expert \\
\midrule
Mean+Ridge
  & $.810 \pm .009$ & $.702 \pm .014$ & $.272 \pm .004$ & $.503 \pm .012$ \\
Attention+Attention
  & $\mathbf{.848 \pm .020}$ & $\mathbf{.728 \pm .025}$ & $.329 \pm .011$ & $.571 \pm .028$ \\
Softmax+Softmax
  & $.810 \pm .013$ & $.585 \pm .010$ & $\mathbf{.497 \pm .017}$ & $.585 \pm .009$ \\
MLP+MLP
  & $.811 \pm .015$ & $.699 \pm .007$ & $.303 \pm .011$ & $\mathbf{.597 \pm .009}$ \\
\bottomrule
\end{tabular}
\caption{Safety- and DV-probe performance across architectures (mean $\pm$ standard deviation over three seeds). DV quality is measured by rank correlation with the continuous delegation value. The best result in each column is bold.}
\label{tab:probe_architecture}
\end{table*}

\begin{figure*}[t]
\centering
\includegraphics[width=\textwidth]{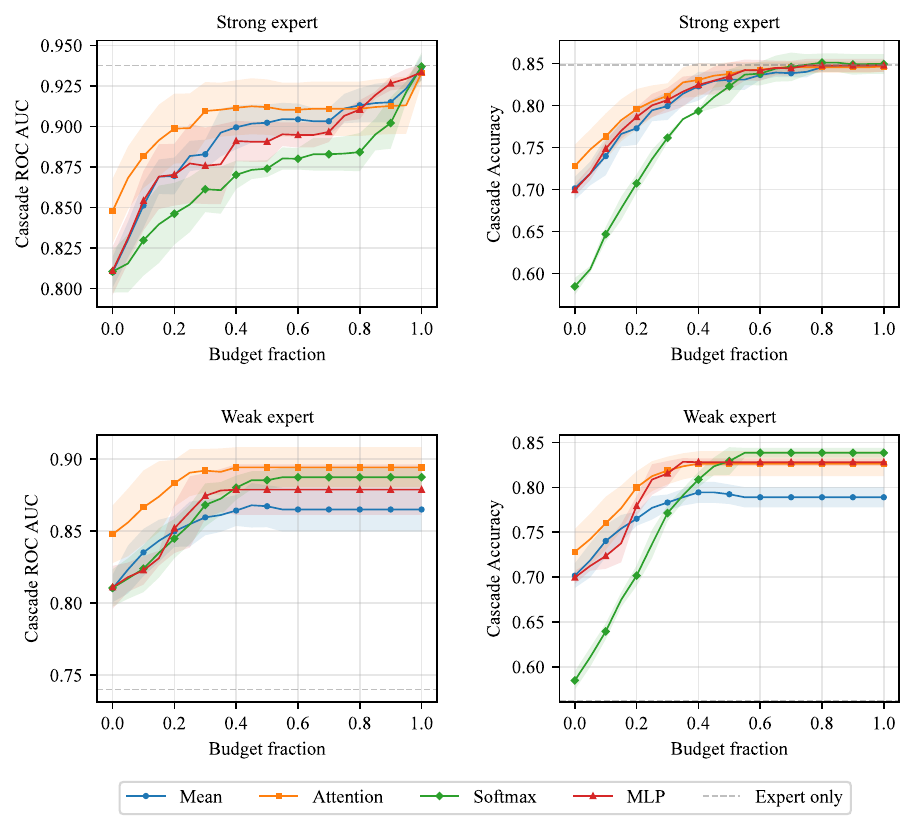}
\caption{Cross-architecture comparison of \theabbrv{} over the complete target-budget sweep at $B\!=\!128$. Rows show the strong and weak experts. Left-column policies optimise AUROC; right-column policies optimise accuracy. Curves and bands are the mean and one standard deviation over three training seeds. Except for Mean, which denotes Mean+Ridge, each label means that the same architecture is used for both probes. The expert-only baseline is shown for reference.}
\label{fig:architecture_ablation}
\end{figure*}

Table~\ref{tab:probe_architecture} shows that learned attention gives the strongest safety probe. Although the highest standalone DV rank correlation varies by expert, Figure~\ref{fig:architecture_ablation} shows that Attention+Attention gives the strongest or comparable absolute cascade performance over most budgets. More importantly, the main qualitative findings persist across architectures: DV-based routing improves on uncertainty-based routing in the useful budget range, and calibrated policies can leave budget unused when further delegation has low empirical value.

\section{Score-Merging Controls}
\label{app:fusion_controls}

\begin{figure*}[t]
\centering
\includegraphics[width=\textwidth]{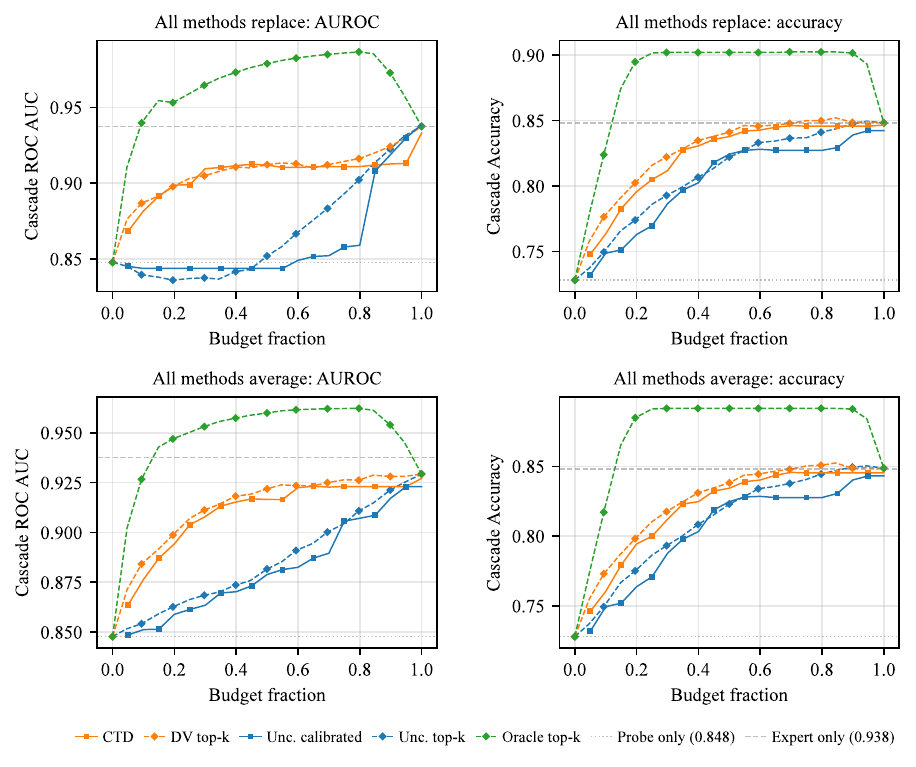}
\caption{Fixed score-merging controls with learned-attention probes and the strong expert. \textbf{Top}: every method overwrites the probe score on delegated inputs. \textbf{Bottom}: every method averages the probe and expert scores. The left panel in each row optimises AUROC for calibrated policies; the right panel optimises accuracy. Curves show means over three seeds.}
\label{fig:fusion_control_strong}
\end{figure*}

\begin{figure*}[t]
\centering
\includegraphics[width=\textwidth]{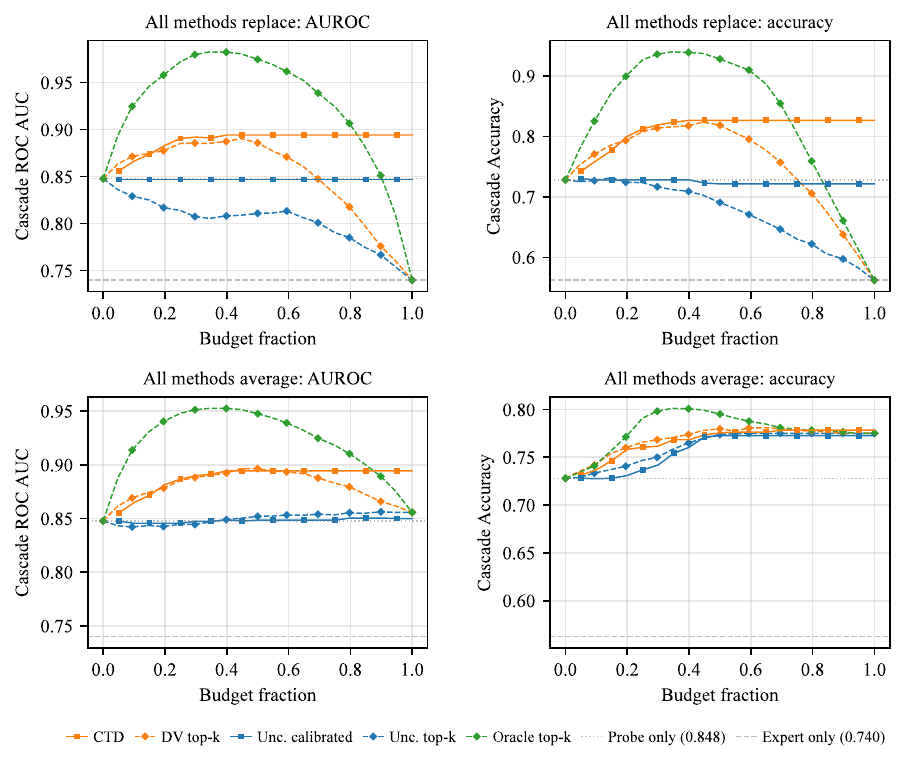}
\caption{Fixed score-merging controls with learned-attention probes and the weak expert. Under overwrite, poorly routed expert calls can directly harm predictions; averaging hedges this effect but attenuates the benefit of selective expert corrections.}
\label{fig:fusion_control_weak}
\end{figure*}

The primary comparison treats the score-merging rule as part of each complete method: DV-based methods overwrite the probe score with the expert score, while the uncertainty methods use their original averaging rule. To separate routing quality from score merging, Figures~\ref{fig:fusion_control_strong} and~\ref{fig:fusion_control_weak} repeat the learned-attention experiment with a common rule for every method.

With a common overwrite rule, uncertainty routing is especially vulnerable to harmful expert overwrite with the weak expert. A common averaging rule protects every method from this failure mode, but also reduces the accuracy gain from selectively overwriting incorrect probe predictions. Under both controls, DV-based routing retains its advantage over uncertainty routing through the low-to-moderate budget range, and \theabbrv{} retains its plateau behaviour. The primary method-specific comparison therefore does not derive its qualitative advantage solely from choosing a different score-merging rule. These controls separate inefficient routing from harmful score merging: averaging may hedge the predictive effect of a poor expert call, but the call still consumes compute.

\section{Delegation-Value Probe Ranking}
\label{app:probe_ranking}

\begin{figure*}[t]
\centering
\includegraphics[width=0.9\textwidth]{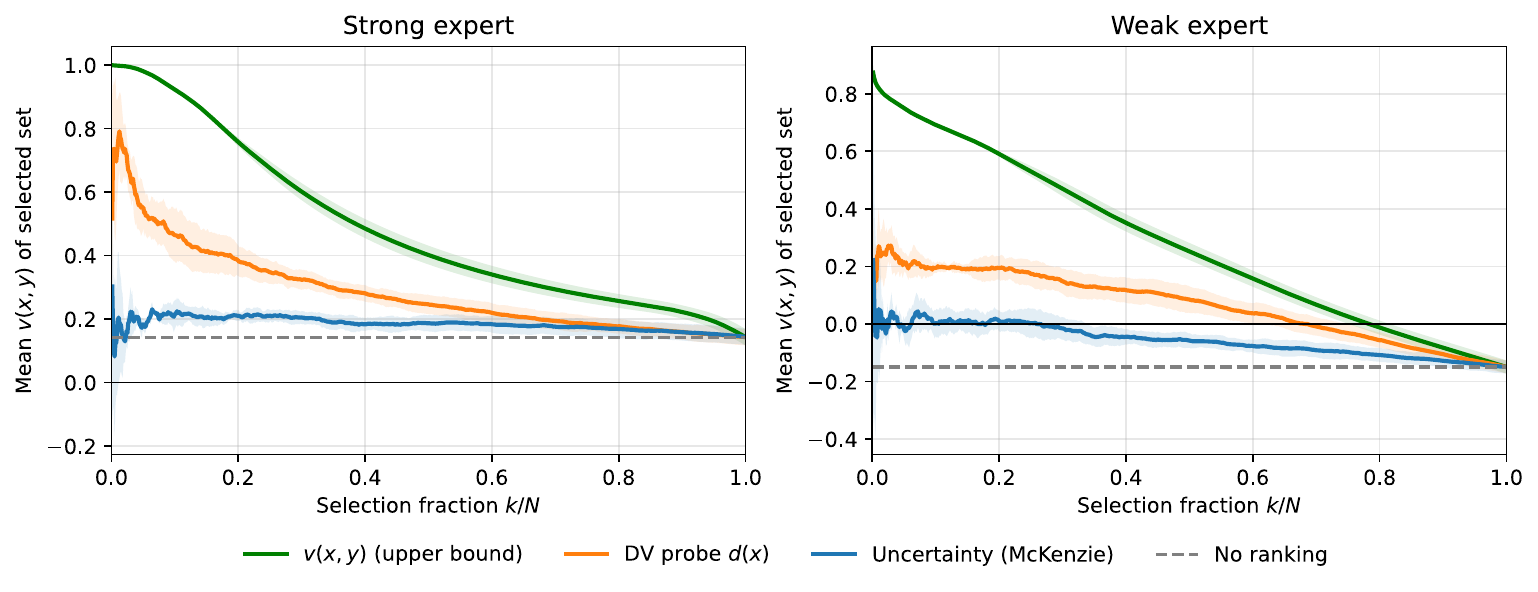}
\caption{Mean $\dv(x,y)$ of the top-$k$ examples ranked by each signal for learned-attention probes, as a function of selection fraction $k/N$. \figleft: strong expert. \figright: weak expert. Lines and bands show the mean and one standard deviation over three seeds. The ground-truth ranking provides an upper bound.}
\label{fig:probe_mean_v_at_k}
\end{figure*}

Figure~\ref{fig:probe_mean_v_at_k} plots the mean delegation value of the top-$k$ examples as ranked by each signal for the Attention+Attention configuration. A higher mean $\dv$ in the selected set means that expert overwrite is more valuable on the selected inputs. Across the selection range and for both weak and strong experts, the DV probe generally ranks above the uncertainty signal. Because this diagnostic concerns the selected inputs rather than the merged score, it exposes low-value expert calls even when averaging makes the final accuracy or AUROC curve appear flat.

\section{Binary Delegation Value formulation}
\label{app:alternative_vx}

An alternative to the continuous delegation value (Equation~6 in the main paper) is a binary indicator that captures only cases where replacing the probe prediction with the expert prediction flips the hard prediction from wrong to right:
\begin{equation}
    \label{eq:dv_binary}
    \dv_{\text{bin}}(x, y) = \mathds{1}\!\big[\text{probe wrong} \;\wedge\; \text{expert correct}\big].
\end{equation}
When \(\dv_{\text{bin}}(x, y) = 1\), expert overwrite corrects the outcome; when \(\dv_{\text{bin}}(x, y) = 0\), overwrite does not help---either both are correct, both are wrong, or the probe is already correct.
This formulation is aligned primarily with accuracy: the oracle ranking by \(\dv_{\text{bin}}\) optimises for flipping wrong-to-right predictions, but treats all such flips equally regardless of the magnitude of score improvement.
In particular, it does not directly optimise AUC, since AUC depends on the \emph{ordering} of scores, not just binary correctness.
The continuous formulation \(\dv(x, y) = P_\expert(y \mid x) - P_\probe(y \mid x)\) captures partial improvements and is better aligned with both AUROC and accuracy. 

\begin{figure*}[t]
\centering
\includegraphics[width=\textwidth]{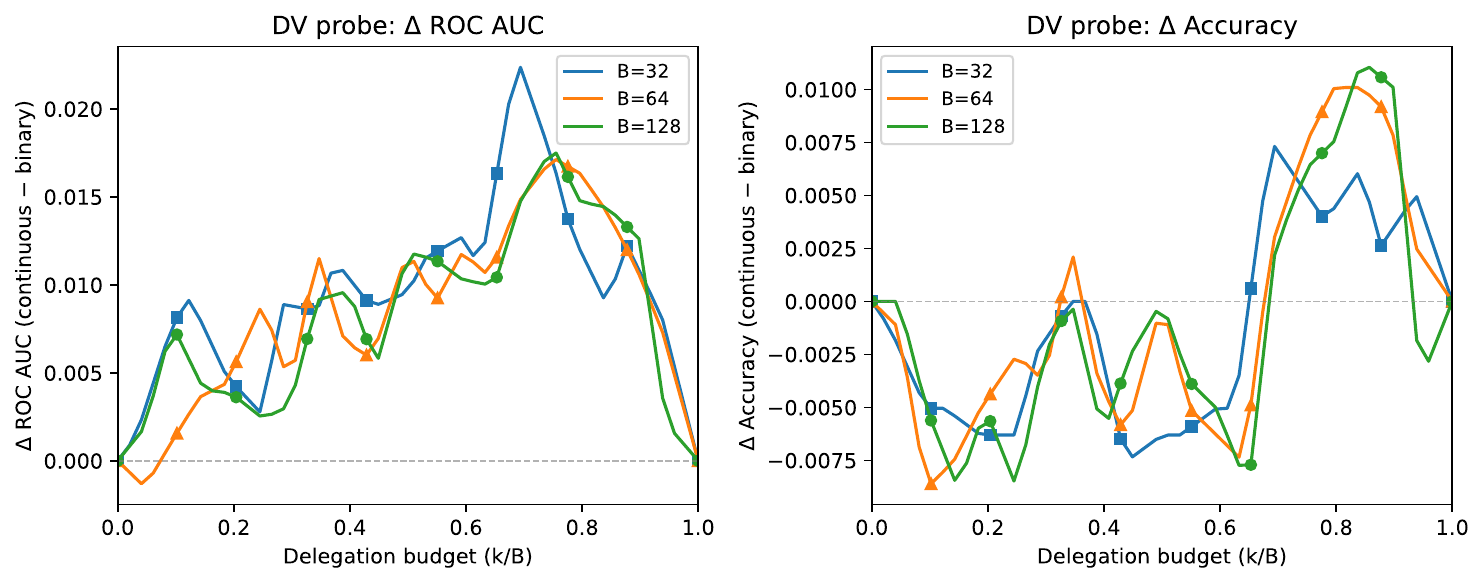}\\[6pt]
\includegraphics[width=\textwidth]{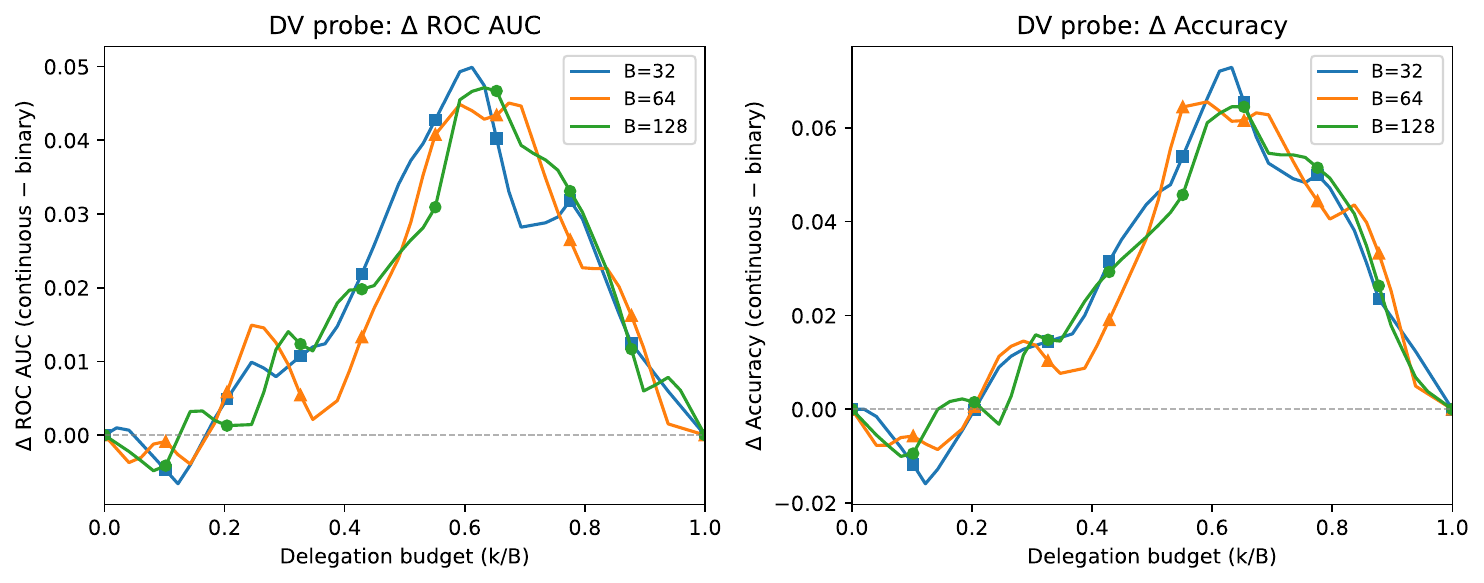}
\caption{Difference in DV top-$k$ performance (continuous $-$ binary) vs.\ delegation budget for three batch sizes. \textbf{Top}: strong expert (Gemma-27B). \textbf{Bottom}: weak expert (Llama-1B). Positive values indicate that the continuous formulation outperforms the binary one. The advantage grows with the budget, consistent with the finer-grained ranking provided by the continuous target.}
\label{fig:dv_delta}
\end{figure*}

Table~\ref{tab:binary_vs_continuous} and Figure~\ref{fig:dv_delta} compare the two formulations empirically using DV top-$k$ delegation at $B=128$. These experiments use the Mean+Ridge configuration to isolate the effect of the target formulation.
With the strong expert, the continuous DV probe matches or exceeds the binary variant in AUC at every budget level (up to $+1.2$ pp at $\budget=0.50$), while accuracy is comparable.
The advantage is more pronounced with the weak expert, where continuous DV improves both AUC ($+2.6$ pp) and accuracy ($+3.8$ pp) at $\budget=0.50$.
The gap widens at higher budgets because the continuous formulation provides a finer-grained ranking: it distinguishes large score improvements from small ones, allowing the top-$k$ set to prioritise examples where expert overwrite helps most.

\begin{table*}[t]
\centering
\small
\setlength{\tabcolsep}{4pt}
\begin{tabular}{@{}l cc cc cc cc@{}}
\toprule
& \multicolumn{4}{c}{Strong expert (Gemma-27B)} & \multicolumn{4}{c}{Weak expert (Llama-1B)} \\
\cmidrule(lr){2-5} \cmidrule(lr){6-9}
& \multicolumn{2}{c}{Binary} & \multicolumn{2}{c}{Continuous} & \multicolumn{2}{c}{Binary} & \multicolumn{2}{c}{Continuous} \\
\cmidrule(lr){2-3} \cmidrule(lr){4-5} \cmidrule(lr){6-7} \cmidrule(lr){8-9}
$\budget$ & AUC & Acc & AUC & Acc & AUC & Acc & AUC & Acc \\
\midrule
10\% & .849 & .757 & .856 & .751 & .834 & .747 & .830 & .737 \\
20\% & .870 & .788 & .874 & .783 & .834 & .757 & .835 & .759 \\
30\% & .880 & .810 & .883 & .808 & .834 & .763 & .849 & .780 \\
50\% & .885 & .830 & .897 & .830 & .825 & .739 & .851 & .777 \\
\bottomrule
\end{tabular}
\caption{DV top-$k$ cascade performance at $B=128$ for binary vs.\ continuous delegation value targets. Continuous DV consistently matches or improves AUC; the gain is largest at high budgets and with the weak expert.}
\label{tab:binary_vs_continuous}
\end{table*}

\section{Additional results on batch size variations}
\label{app:batch_size_variations}

\begin{figure*}[t]
\centering
\includegraphics[width=\textwidth]{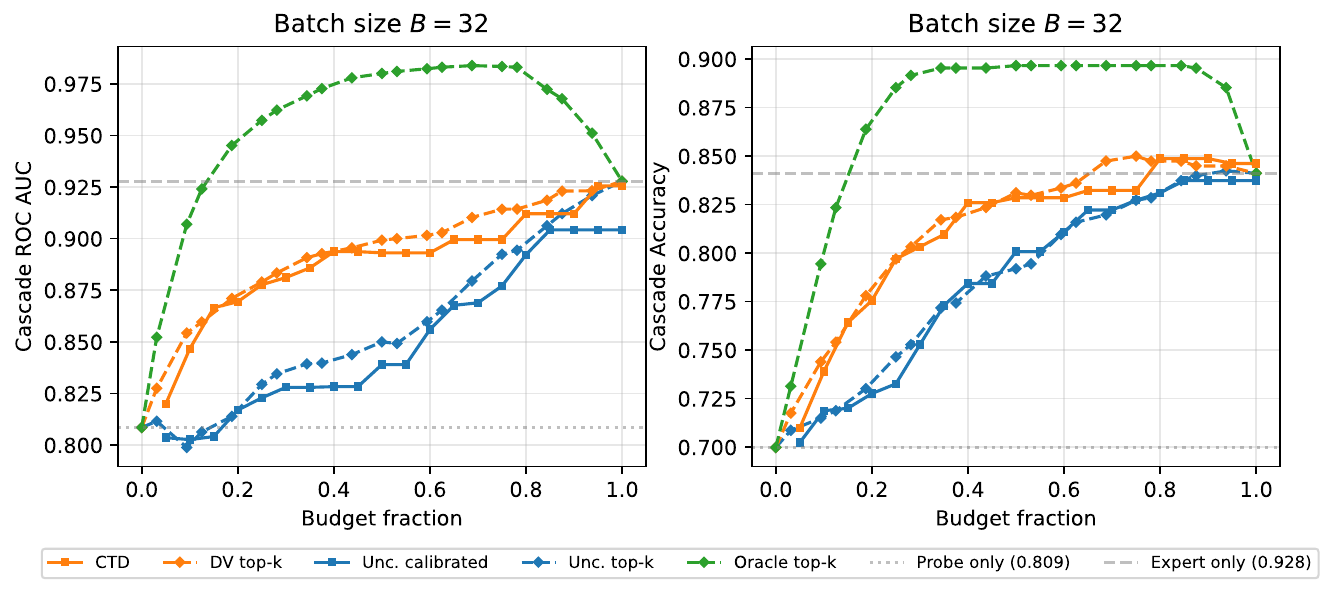}\\[6pt]
\includegraphics[width=\textwidth]{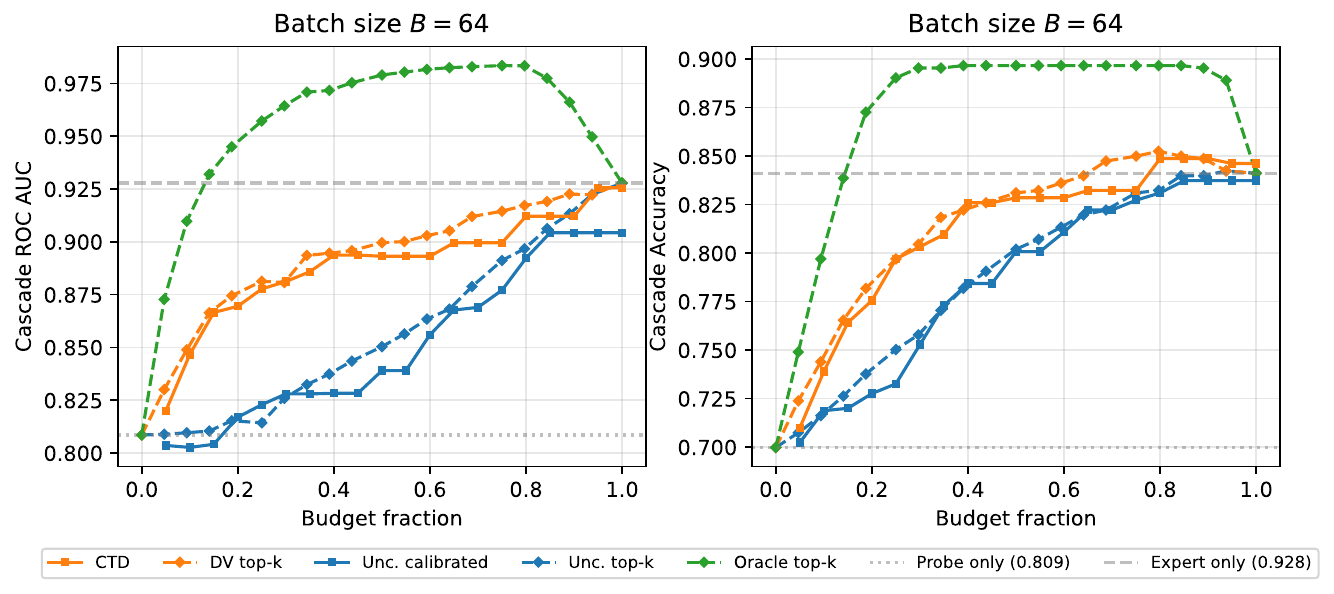}
\caption{Mean+Ridge configuration with the strong expert (Gemma-3-27B-IT): cascade performance at $B\!=\!32$ (top) and $B\!=\!64$ (bottom).}
\label{fig:batch_strong}
\end{figure*}

\begin{figure*}[t]
\centering
\includegraphics[width=\textwidth]{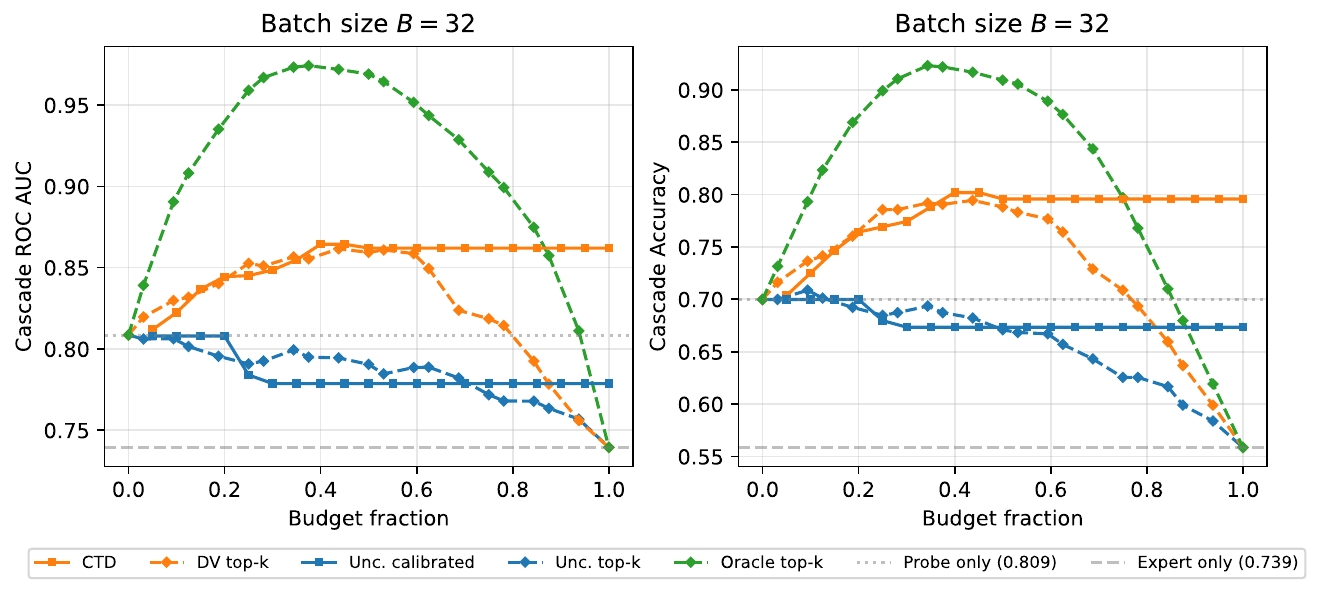}\\[6pt]
\includegraphics[width=\textwidth]{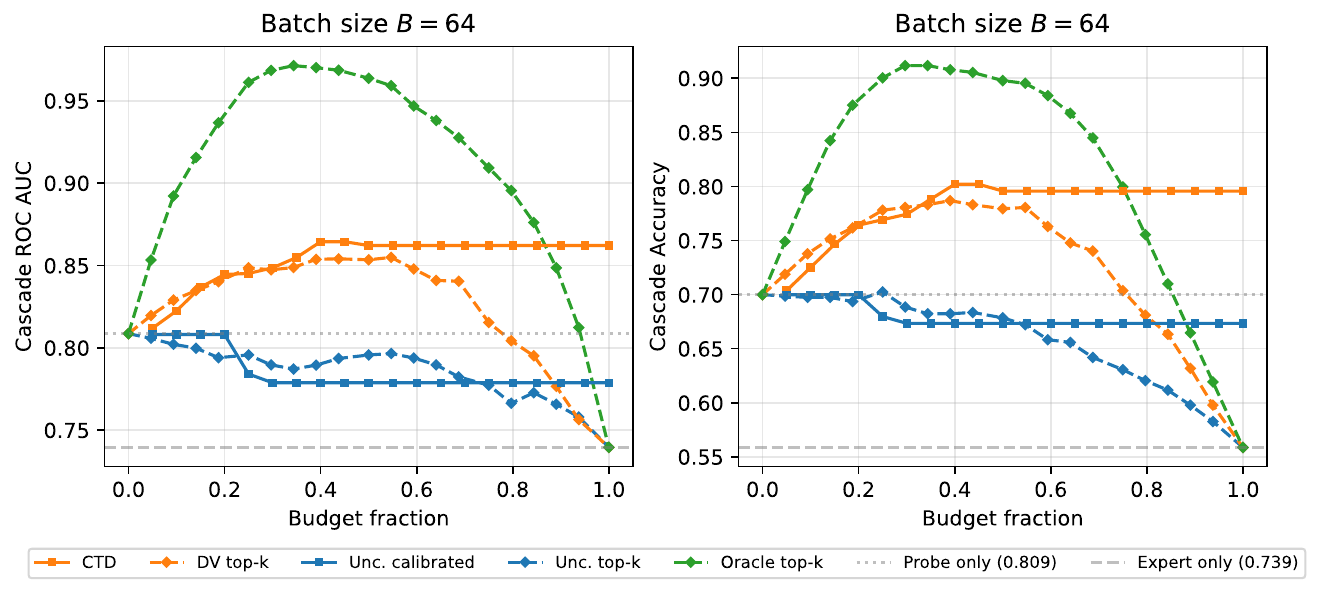}
\caption{Mean+Ridge configuration with the weak expert (Llama-3.2-1B-Instruct): cascade performance at $B\!=\!32$ (top) and $B\!=\!64$ (bottom).}
\label{fig:batch_weak}
\end{figure*}

Figures~\ref{fig:batch_strong} and~\ref{fig:batch_weak} show cascade performance at batch sizes $B\!=\!32$ and $B\!=\!64$ for the Mean+Ridge configuration, complementing the architecture ablation at $B\!=\!128$.
All qualitative findings hold across batch sizes: DV top-$k$ dominates Unc.\ top-$k$, and the calibrated threshold variants plateau beyond the effective delegation capacity instead of degrading.

\section{Group Analysis Across Seeds}
\label{app:group_analysis}

The main-paper group table averages the learned-attention results over three seeds. With the strong expert, the mean value of the \theabbrv{}-delegated subset is positive for every group in every seed. With the weak expert, it is positive in every seed for Anthropic~HH, MTS-Dialog, and ToolACE. MTSamples is the least stable group: its delegated-subset mean is negative for one seed and positive for two, yielding a positive three-seed mean of \(+0.057\). Thus the central result---that DV routing can recover positive-value subsets within groups where the expert score is worse than the probe score on average---is consistent across groups and largely stable across initialisation.

\section{Empirical validation of budget guarantees}
\label{app:budget_coverage}

\begin{figure*}[t]
\centering
\includegraphics[width=0.48\textwidth]{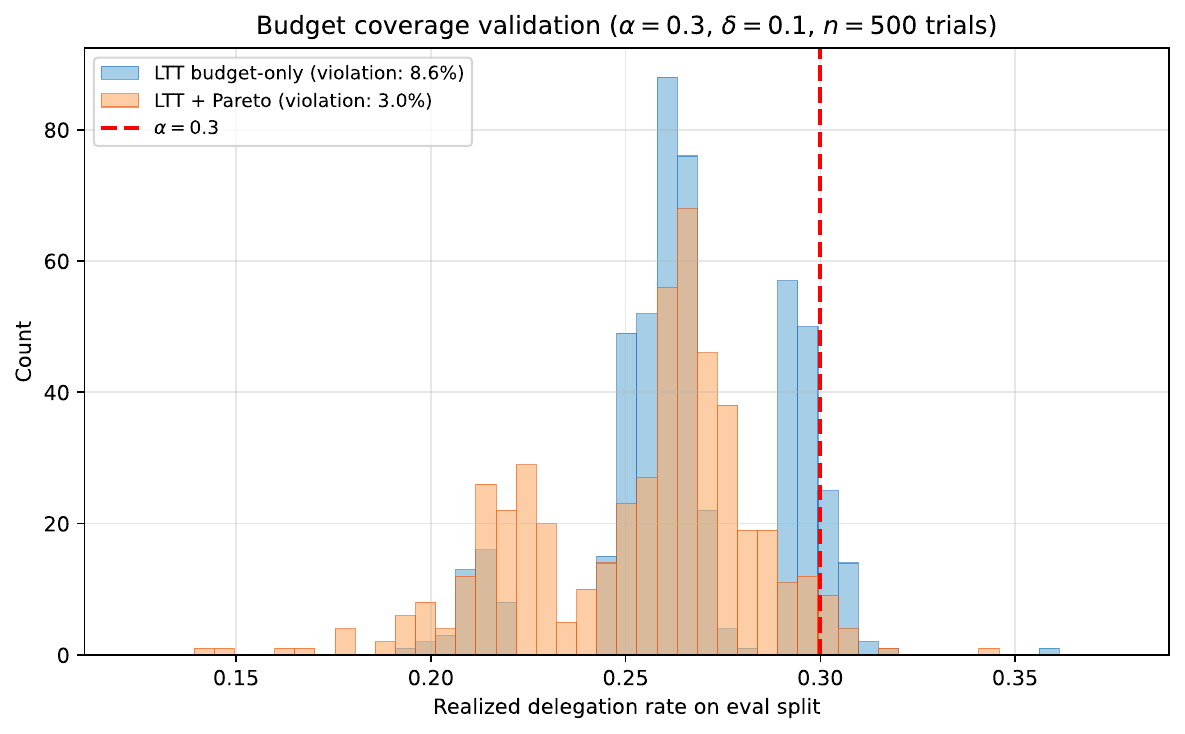}\hfill
\includegraphics[width=0.48\textwidth]{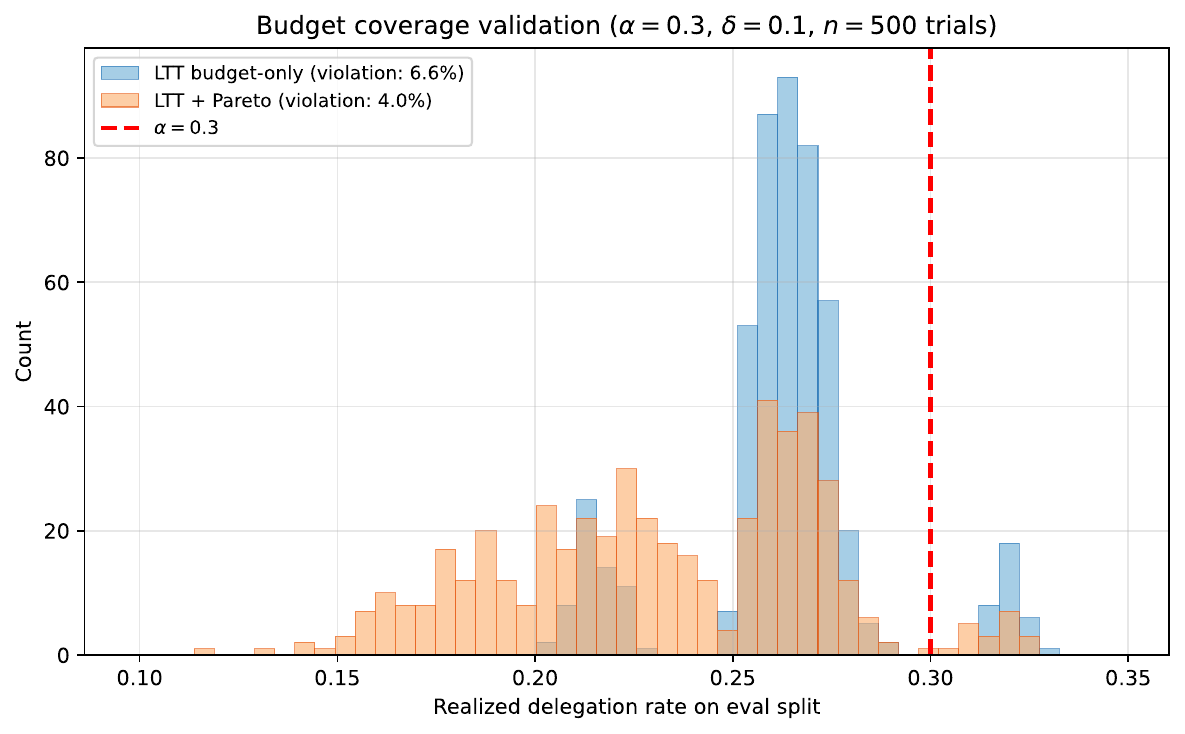}
\caption{Distribution of realised delegation rates across 500 random calibration splits (\(\budget = 0.3\), \(\budgetdelta = 0.1\)). \textbf{Left}: strong expert (Gemma-3-27B-IT). Budget-only LTT violates at 8.6\%, close to \(\budgetdelta\); Pareto LTT is more conservative at 3.0\%. \textbf{Right}: weak expert (Llama-3.2-1B-Instruct). Both methods stay below \(\budgetdelta\) (6.6\% and 4.0\%), with the Pareto filter adapting its conservatism to the limited effective delegation capacity.}
\label{fig:coverage}
\end{figure*}

We empirically validate the PAC budget guarantee \(\Pr(\Rbudget(\lambda^*) \leq \budget) \geq 1-\budgetdelta\) by repeating the calibration procedure across 500 independent random splits of the test data into calibration and evaluation subsets. For each trial, we calibrate the threshold on the calibration split (using either budget-only LTT or Pareto-filtered LTT) and measure the realised delegation rate on the held-out evaluation split. This validation uses the Mean+Ridge configuration; the guarantee itself is independent of the probe architecture. Figure~\ref{fig:coverage} shows the results at \(\budget = 0.3\) and \(\budgetdelta = 0.1\). Budget-only LTT is near-tight: its violation rate (8.6\% with the strong expert, 6.6\% with the weak) closely tracks~\(\budgetdelta\). Pareto-filtered LTT is more conservative (3.0\% and 4.0\% respectively), because the Pareto filter restricts the candidate set by discarding thresholds that would hurt performance. With a weak expert, positive-value overwrite opportunities are rare, so the Pareto filter prunes more aggressively and realised rates fall further below~\(\budget\).

\section{Performance Control with Accuracy Risk}
\label{app:accuracy_guarantees}

We repeat the performance-risk validation from the main paper for all four safety/DV probe architectures. We use \(\performance=0.20\) with the strong expert and \(\performance=0.22\) with the weak expert, corresponding to minimum accuracies of 0.80 and 0.78, with confidence \(1-\performancedelta=0.90\). In each of 500 random splits, \theabbrv{}'s LTT selector and the plug-in empirical selector use the same candidate thresholds, Pareto filter, optimisation data, and hypothesis-testing data. The empirical selector replaces the binomial test with the naive requirement that observed calibration error not exceed \(\performance\).

\begin{figure*}[t]
\centering
\includegraphics[width=\textwidth]{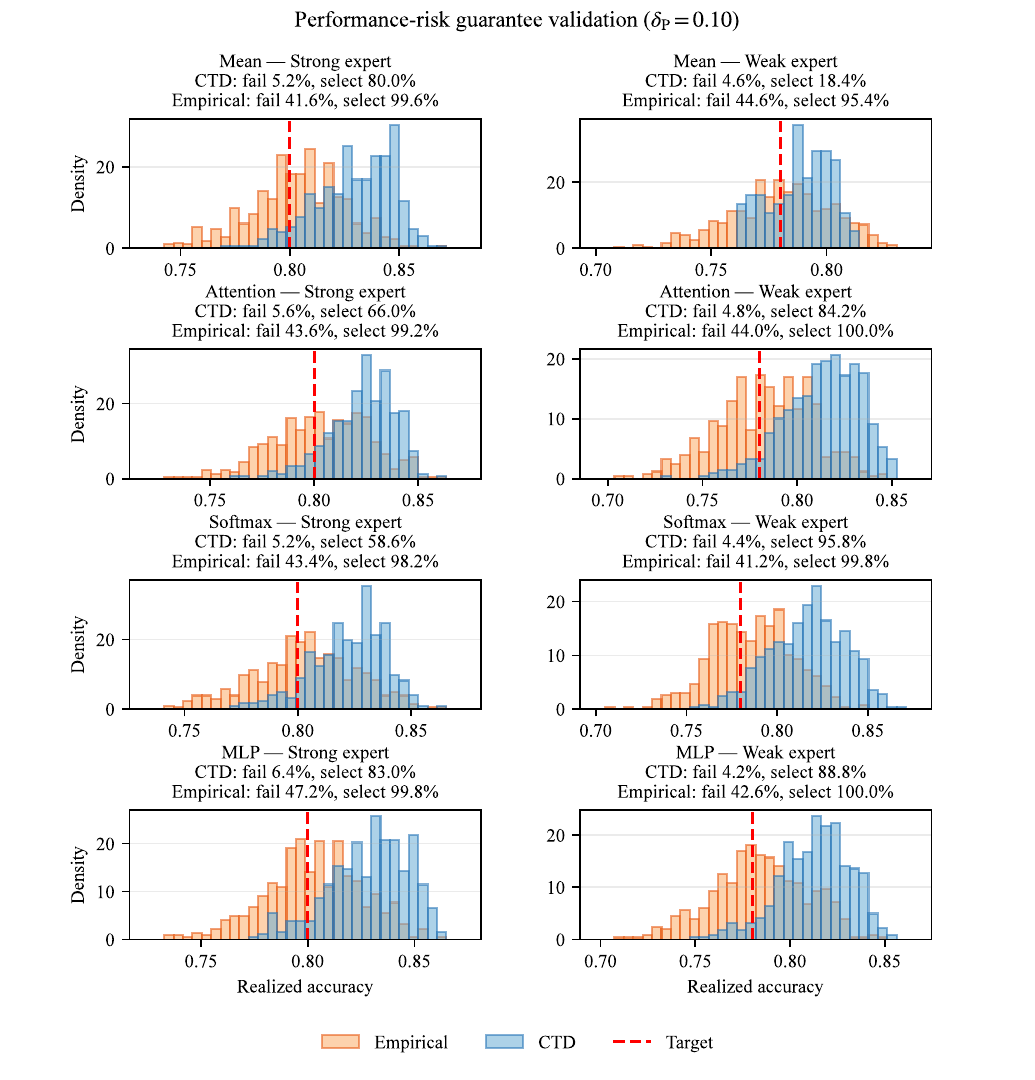}
\caption{Performance-risk guarantee validation across probe architectures over 500 random splits (\(\performancedelta=0.10\)). Rows show probe architectures; columns show strong and weak experts. Dashed red lines mark accuracy \(1-\performance\). Histograms condition on selection, while titles report unconditional violation and selection rates.}
\label{fig:accuracy_guarantee_all_architectures}
\end{figure*}

Figure~\ref{fig:accuracy_guarantee_all_architectures} shows the held-out distributions. Across architectures, \theabbrv{}'s unconditional violation rate ranges from 4.2\% to 6.4\%, below the target \(\performancedelta=10\%\), whereas empirical calibration violates in 41.2\%--47.2\% of trials. Selection rates vary because the attainable target and statistical evidence depend on the learned probe and expert. This is especially visible for Mean+Ridge with the weak expert, where \theabbrv{} selects in only 18.4\% of trials. Reporting selection alongside violations distinguishes reliable non-selection from deploying a policy without adequate evidence.

\begin{figure*}[t]
\centering
\includegraphics[width=0.9\textwidth]{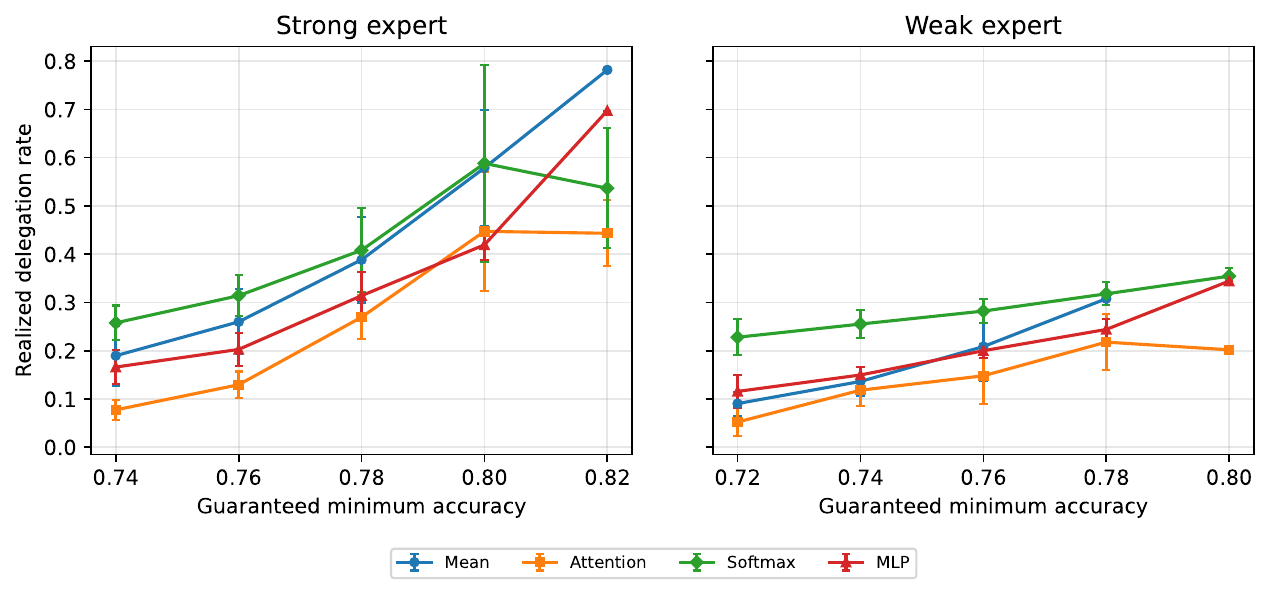}
\caption{Realised delegation rate versus guaranteed minimum accuracy for the four probe architectures. \textbf{Left}: strong expert. \textbf{Right}: weak expert. Points and error bars show the mean and one standard deviation over three training seeds among certified policies.}
\label{fig:accuracy_budget_tradeoff}
\end{figure*}

Figure~\ref{fig:accuracy_budget_tradeoff} complements the fixed-target validation by sweeping the guaranteed minimum accuracy. More demanding guarantees generally require more expert calls, although finite samples and non-monotone cascade accuracy can produce small local reversals. Learned attention requires the least delegation over most of the sweep, particularly with the weak expert. Error bars show variation over training seeds; points are omitted when no policy is certified.

\section{Sensitivity to the Pareto Objective}
\label{app:auroc_risk}

\begin{figure*}[t]
\centering
\includegraphics[width=\textwidth]{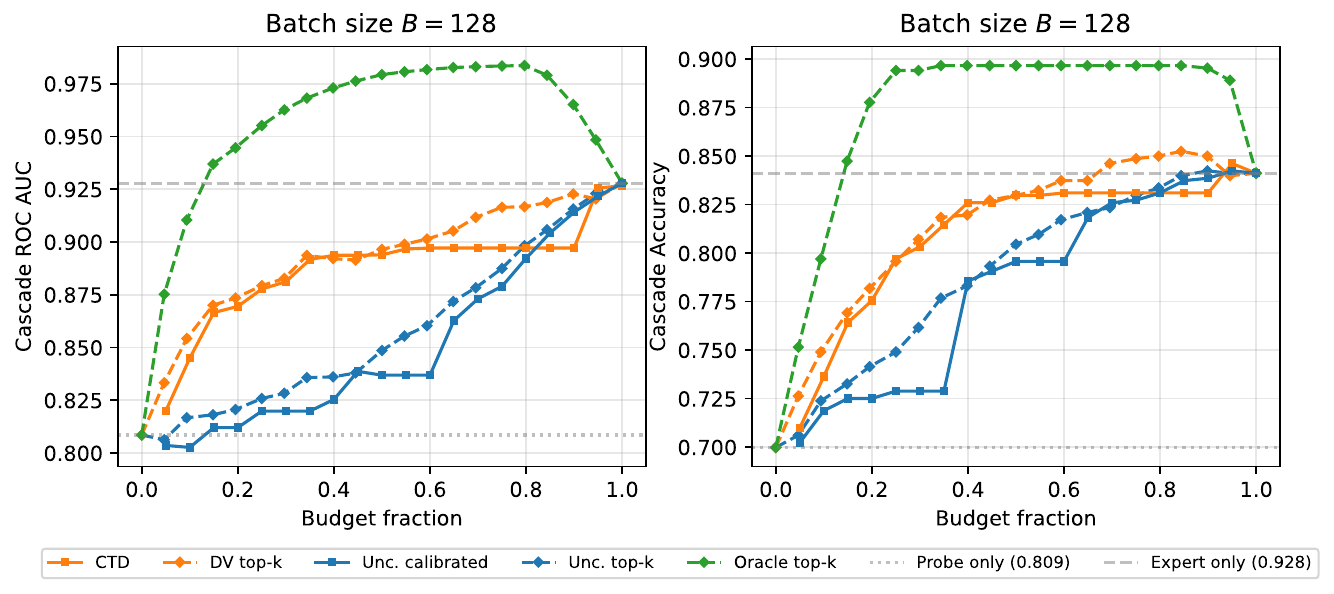}
\caption{Mean+Ridge cascade performance with AUROC error as the Pareto-testing objective in both panels (strong expert, $B\!=\!128$).}
\label{fig:auroc_strong}
\end{figure*}

\begin{figure*}[t]
\centering
\includegraphics[width=\textwidth]{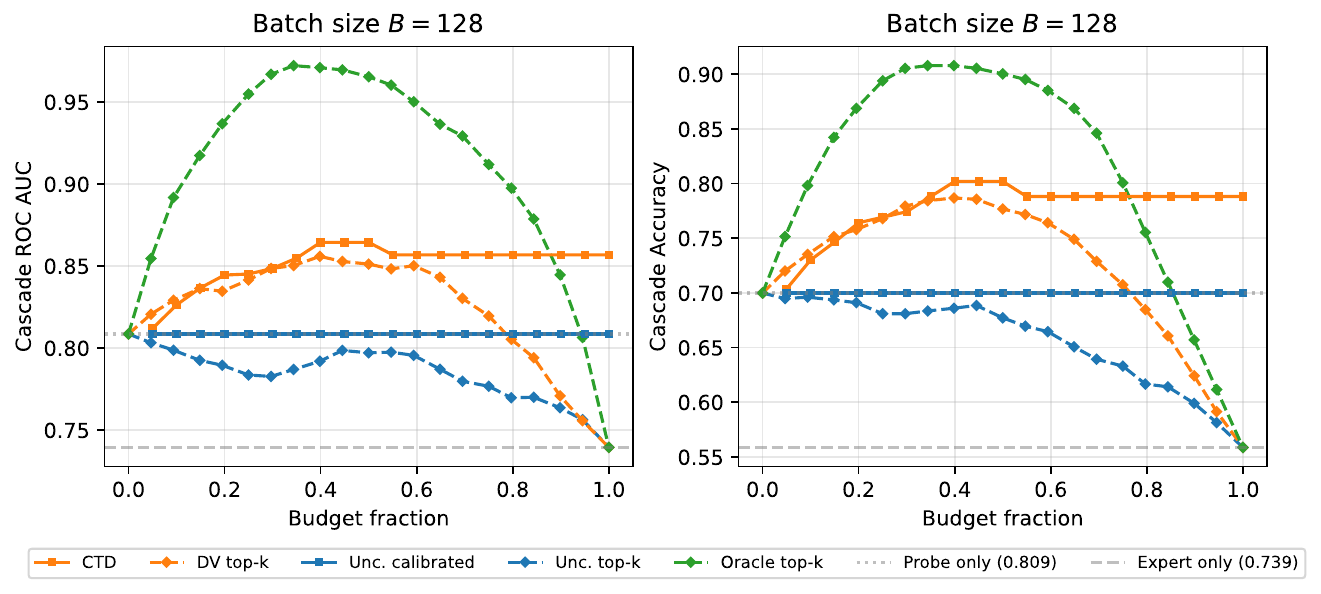}
\caption{Mean+Ridge cascade performance with AUROC error as the Pareto-testing objective in both panels (weak expert, $B\!=\!128$).}
\label{fig:auroc_weak}
\end{figure*}

In the main composite figures, the calibrated policy is selected using the metric shown in each panel: AUROC error for the AUROC panel and accuracy error for the accuracy panel.
To expose the effect of using a single objective, Figures~\ref{fig:auroc_strong} and~\ref{fig:auroc_weak} instead show both metrics when AUROC error ($1-\mathrm{AUROC}$) is used for Pareto filtering in both panels.
These experiments use the Mean+Ridge configuration. The selected thresholds can change with the objective, but the qualitative findings are unchanged: the DV delegation signal improves on uncertainty, and the calibrated threshold variants plateau beyond the effective delegation capacity.

\end{document}